\documentclass[sigconf]{aamas}

\def\Real{\mathbb{R}}
\usepackage{color}

\usepackage{amsmath,amssymb}
\usepackage{amsthm,bm}
\usepackage{amsmath}
\usepackage{cases}

\newtheorem{theorem}{Theorem}
\newtheorem{assumption}{Assumption}

\newtheorem{lemma}{Lemma}

\newenvironment{myproof}{{\noindent\it Proof. }}{\hfill $\square$ \par}

\definecolor{mygreen}{rgb}{0.10,0.50,0.10}

\DeclareMathOperator*{\argmax}{arg\,max}

\usepackage{comment}

\usepackage{subfigure}

\usepackage{algorithm}
\usepackage{algorithmic}
\usepackage{float}
\usepackage{threeparttable}

\usepackage{balance} 

\setcopyright{ifaamas}
\acmConference[AAMAS '24]{Proc.\@ of the 23rd International Conference
on Autonomous Agents and Multiagent Systems (AAMAS 2024)}{May 6 -- 10, 2024}
{Auckland, New Zealand}{N.~Alechina, V.~Dignum, M.~Dastani, J.S.~Sichman (eds.)}
\copyrightyear{2024}
\acmYear{2024}
\acmDOI{}
\acmPrice{}
\acmISBN{}

\acmSubmissionID{505}

\title{Adaptive Primal-Dual Method for Safe Reinforcement Learning}

\author{Weiqin Chen}
\affiliation{
  \institution{Rensselaer Polytechnic Institute}
  \city{Troy, NY}
  \country{USA}}
\email{chenw18@rpi.edu}

\author{James Onyejizu}
\affiliation{
  \institution{Rensselaer Polytechnic Institute}
  \city{Troy, NY}
  \country{USA}}
\email{onyejj@rpi.edu}

\author{Long Vu}
\affiliation{
  \institution{IBM T.J. Watson Research Center}
  \city{Yorktown Heights, NY}
  \country{USA}}
\email{lhvu@us.ibm.com}

\author{Lan Hoang}
\affiliation{
  \institution{IBM Research}
  \city{Daresbury, Warrington}
  \country{UK}}
\email{lan.hoang@ibm.com}

\author{Dharmashankar Subramanian}
\affiliation{
  \institution{IBM T.J. Watson Research Center}
  \city{Yorktown Heights, NY}
  \country{USA}}
\email{dharmash@us.ibm.com}

\author{Koushik Kar}
\affiliation{
  \institution{Rensselaer Polytechnic Institute}
  \city{Troy, NY}
  \country{USA}}
\email{kark@rpi.edu}

\author{Sandipan Mishra}
\affiliation{
  \institution{Rensselaer Polytechnic Institute}
  \city{Troy, NY}
  \country{USA}}
\email{mishrs2@rpi.edu}

\author{Santiago Paternain}
\affiliation{
  \institution{Rensselaer Polytechnic Institute}
  \city{Troy, NY}
  \country{USA}}
\email{paters@rpi.edu}

\begin{abstract}
Primal-dual methods have a natural application in Safe Reinforcement Learning (SRL), posed as a constrained policy optimization problem. In practice however, applying primal-dual methods to SRL is challenging, due to the inter-dependency of the learning rate (LR) and Lagrangian multipliers (dual variables) each time an embedded unconstrained RL problem is solved. In this paper, we propose, analyze and evaluate adaptive primal-dual (APD) methods for SRL, where two adaptive LRs are adjusted to the Lagrangian multipliers so as to optimize the policy in each iteration. We theoretically establish the convergence, optimality and feasibility of the APD algorithm. Finally, we conduct numerical evaluation of the practical APD algorithm with four well-known environments in Bullet-Safey-Gym employing two state-of-the-art SRL algorithms: PPO-Lagrangian and DDPG-Lagrangian. All experiments show that the practical APD algorithm outperforms (or achieves comparable performance) and attains more stable training than the constant LR cases. Additionally, we substantiate the robustness of selecting the two adaptive LRs by empirical evidence.
\end{abstract}


\newcommand{\BibTeX}{\rm B\kern-.05em{\sc i\kern-.025em b}\kern-.08em\TeX}

\begin{document}


\pagestyle{fancy}
\fancyhead{}


\maketitle


\section{Introduction}\label{section_intro}
Reinforcement learning (RL) has a rich history of solving a wide range of decision-making problems. Recently, RL has succeeded in training large language models such as ChatGPT~\cite{schulman2022chatgpt}, playing video games at superhuman level~\cite{mnih2013playing,mnih2015human,mnih2016asynchronous}, mastering Go~\cite{silver2017mastering,silver2018general}, and manipulating robotics~\cite{nguyen2019review,levine2016end}. RL problems, in general, are formulated as Markov Decision Processes (MDPs). In this work, we are interested in conditions where the underlying dynamics are unknown, the optimal policy thus needs to be learned from data (samples). The goal for an agent is to explore the environment so that it is able to maximize the expected cumulative reward.

Nevertheless, classical RL techniques might lead to risky/unsafe actions~\cite{garcia2015comprehensive,gu2022review,chen2023policy}, if they are only concerned with the reward. Therefore, safety constitutes a foundational aspect in realistic domains or physical entities. Specifically, in the realm of robot navigation, ensuring collision avoidance~\cite{kahn2018self,zhu2021deep} is essential for their proper functioning, and to ensure the preservation of human safety in the vicinity. Taking into account the safety requirements motivates the development of policy optimization under safety guarantees~\cite{geibel2006reinforcement,kadota2006discounted,chow2017risk}.

A common approach is to employ the framework of Constrained MDPs (CMDPs)~\cite{altman1999constrained}
where auxiliary costs (analogous to reward) are considered in the constraints. This framework has gained widespread adoption for inducing safe behaviors~\cite{achiam2017constrained, yang2020projection,zhang2020first,chow2017risk,tessler2018reward,paternain2019constrained}. We briefly introduce these work below.

\subsection{Related Work}
The state-of-the-art algorithms for solving CMDPs commonly include two types of methods: primal methods and primal-dual methods.

\subsubsection{Primal Methods.}
\cite{achiam2017constrained} 
develops a constrained policy optimization (CPO) algorithm for SRL that searches the feasible policy within the confines of the trust region while guaranteeing a monotonic performance improvement as well as constraint satisfaction. Projection-based constrained policy optimization (PCPO)~\cite{yang2020projection} employs TRPO~\cite{schulman2015trust}, the cutting-edge unconstrained RL algorithm first, and then projects the policy back into the feasible set. Nevertheless, both CPO and PCPO suffer from the approximation error and expensive computation of the inversion of high-dimensional Hessian matrices. To address these issues, \cite{zhang2020first} proposes first-order constrained optimization in policy space (FOCOPS), which optimizes the constrained policy in the non-parametric space, and subsequently derives the first-order gradients of the $\ell_2$ loss function in the parameterization space. Another alternative is the penalized proximal policy optimization (P3O)~\cite{zhang2022penalized}, which solves the constrained policy iteration based on the minimization of an equivalent unconstrained problem. However, both FOCOPS and P3O introduce more auxiliary parameters that need to be learned.

\subsubsection{Primal-Dual Methods.}
Lagrangian-based primal-dual algorithms like primal-dual optimization (PDO) \cite{chow2017risk} and reward constrained policy optimization (RCPO)~\cite{tessler2018reward} have succeeded in solving CMDP optimization problems. Nevertheless, the convergence guarantees of these algorithms are limited to either local (locally optimal policies or stationary-point)~\cite{bhatnagar2012online, chow2017risk, tessler2018reward} or asymptotic scenarios~\cite{borkar2005actor}. \cite{paternain2019constrained} establishes the analysis on the duality gap for CMDPs within the policy space and provides a provably dual descent algorithm under the assumption of access to a non-convex optimization oracle. However, obtaining the solution to a primal non-convex problem remains an issue, thus lacking global convergence guarantees. To tackle these challenges, \cite{ding2020natural} proposes a natural policy gradient primal-dual (NPG-PD) method that
achieves non-asymptotic global convergence with sublinear rates regarding both the optimality gap and the constraint violation. \cite{ding2021provably} provides the first provably efficient online policy optimization algorithm--optimistic primal-dual proximal policy optimization (OPDOP) and establishes the bounds on the regret and constraint violation. \cite{chen2023probabilistic} proposes a safe primal-dual (SPG) algorithm that is used to solve a CMDP problem with probabilistic safety constraints. Nonetheless, achieving satisfactory performance with the existing primal-dual methods remains challenging, mainly due to their sensitivity to hyper-parameters such as learning rates.

\subsection{Main Contribution}
This paper addresses a core challenge in applying primal-dual methods to SRL problems, the inter-dependence of the primal \textit{Learning Rate (LR)} and \textit{Lagrangian Multiplier (LM)} (dual variable) parameters 
in the primal-dual method. 
We provide analytical expressions (bounds) of the amount of progress made in each step of the primal-dual algorithm, based on which we develop two adaptive LR choices that optimize these bounds. The two LR choices have an inverse-linear and inverse-quadratic dependence on the LM, and are incorporated into the proposed adaptive primal-dual (APD) algorithm. 
We provide theoretical analyses of the convergence, return optimality, and constraint feasibility of the APD algorithm. Finally, we numerically evaluate the practical version of APD (PAPD) algorithm using four environments in the Bullet-Safety-Gym\cite{gronauer2022bullet}, and compare the performance of PAPD with constant-LR solutions using two state-of-the-art constrained RL methods: PPO-Lagrangian (PPOL) and DDPG-Lagrangian (DDPGL). We also numerically demonstrate the robustness of PAPD algorithm with respect to certain key parameter choices.


\section{Safe Reinforcement Learning}
MDPs are defined by a tuple ($S, A, R, \mathbb{P}, \mathcal{U}, \gamma$)~\cite{sutton2018reinforcement}, where $S$ is the state space, $A$ is the action space, $R: S \times A \times S \to \Real$ is the reward function describing the quality of the decision. For any $\hat{S} \subset S, s_t \in S, a_t \in A, t\in \{0, 1, \cdots \}$, $\mathbb{P}(s_{t+1}\in\hat{S} | s_t, a_t)$ (i.e., the probability of $s_{t+1}$ being in $\hat{S}$ given $s_t$ and $a_t$) is the transition probability describing the dynamics of the system, $\mathcal{U}(\hat{S}):= \mathbb{P}(s \in \hat{S)}$ is the starting state distribution, and $\gamma$ is the discount factor. Note that a table of notations is provided in the supplementary material, which serves to facilitate tracking and enhance comprehension for the readers.

Consider a parameterization space $\Theta$ and a probability density function $P(\cdot)$. Given $\theta \in \Theta$, a parameterized stationary \emph{policy} $\pi_\theta : S \to P(A)$ maps states to probability distributions over the set of actions, and $\pi_\theta(a|s)$ indicates the probability density of drawing action $a \in A$ in the corresponding state $s \in S$. Common parameterizations include neural networks (NN) and Radial Basis Functions (RBFs). In this work, we are particularly interested in situations where the state transition distributions are unknown, and thus the policies need to be computed through interacting with the environment. 

In the context of MDPs, the objective is to find an optimal parameter that maximizes the expected
discounted return
\begin{align}\label{eqn_performance_measure}
    J_R (\pi_\theta) = \mathbb{E}_{\tau \sim \pi_\theta} \left[ \sum_{t=0}^{\infty} \gamma^t R(s_t, a_t, s_{t+1}) \right],
\end{align}
where $\tau = (s_0, a_0, s_1, a_1, \cdots)$ denotes a sample trajectory. Given fixed initial state distribution $s_0 \sim \mathcal{U}$ and transition distribution $s_{t+1} \sim \mathbb{P}(\cdot|s_t, a_t)$, let us define a shorthand $\tau \sim \pi_\theta$ indicating that the distribution over trajectories depends on the policy through $a_t \sim \pi_\theta(\cdot | s_t)$.

CMDPs~\cite{altman1999constrained} impose additional constraints on the allowable policies. More concretely, auxiliary cost functions $C_i: S \times A \times S \to \Real, i=1, 2, \cdots, m$ are introduced. We make the following assumption about the costs. 

\begin{assumption}\label{assumption_bound_cost}
    Consider $B>0$ and $C=[C_1, \cdots, C_m]^T$, assume we have bounded costs such that $||C|| \leq B$.
\end{assumption}

Analogous to the expected discounted return, define the expected discounted cost as
\begin{align}\label{eqn_performance_measure_cost}
    J_{C_i} (\pi_\theta) = \mathbb{E}_{\tau \sim \pi_\theta} \left[ \sum_{t=0}^{\infty} \gamma^t C_i(s_t, a_t, s_{t+1}) \right].
\end{align}

Denote by $d_i, i=1, 2, \cdots, m$ the desired threshold for the expected discounted cost. Accordingly, SRL problems can be formulated as
\begin{align}\label{eqn_SRL_problem}
    &\theta^* \in \argmax_{\theta \in \Theta} \, J_R(\pi_\theta) \nonumber \\
    &\quad\quad\,\,\text{s.t.} \quad J_{C_i}(\pi_\theta) \leq d_i, \, i=1, 2, \cdots, m.
\end{align}
The SRL problem posed in \eqref{eqn_SRL_problem} can be solved by applying gradient-based methods on a regularized objective function~\cite{censor1977pareto} or using primal-dual methods~\cite{chow2017risk,chen2023probabilistic} to achieve local optimal solutions. The performance of the first type of method can heavily depend on the choice of the regularization parameter, and this choice is problem dependent for which there is no easy method. In general, for a given choice of regularization parameter, the first type of method may not produce optimal results. In this work, therefore, we focus on the primal-dual method, where the regularization parameter is the LM (dual variable) that is progressively updated. We introduce this method in the next section and motivate the need for an adaptive LR in the primal update step.


\section{Adaptive Primal-Dual Algorithm}

\subsection{Motivation}
%

Primal-dual methods rely on iterative training of the policy parameters $\theta$ to minimize the Lagrangian of \eqref{eqn_SRL_problem}:
\begin{align}\label{eqn_SRL_problem_lagrangian}
    \mathcal{L}(\theta, \lambda) \dot{=} -J_R(\pi_{\theta}) + \lambda^T (J_{C}(\pi_\theta) - {\bf d}),
\end{align}
where $J_{C}(\cdot) = [J_{C_1}(\cdot), J_{C_2}(\cdot), \cdots, J_{C_m}(\cdot)]^T, \, {\bf d} = [d_1, d_2, \cdots, d_m]^T$, and $\lambda \in \Real^m$ is the LM.

\begin{figure}
\centering
\subfigure{
\begin{minipage}[t]{0.22\columnwidth}
\centering
\includegraphics[width=1\textwidth]{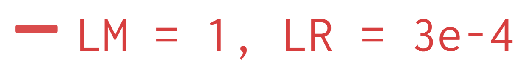}
\end{minipage}
}
\subfigure{
\begin{minipage}[t]{0.22\columnwidth}
\centering
\includegraphics[width=1\textwidth]{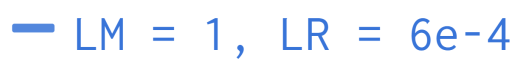}
\end{minipage}
}
\subfigure{
\begin{minipage}[t]{0.22\columnwidth}
\centering
\includegraphics[width=1\textwidth]{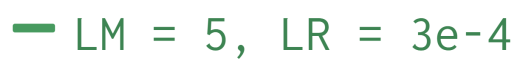}
\end{minipage}
}
\subfigure{
\begin{minipage}[t]{0.22\columnwidth}
\centering
\includegraphics[width=1\textwidth]{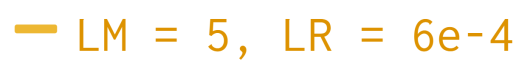}
\end{minipage}
}
\setcounter{subfigure}{0}
\subfigure[Return]{
\begin{minipage}[t]{0.47\columnwidth}
\centering
\includegraphics[width=1.1\textwidth]{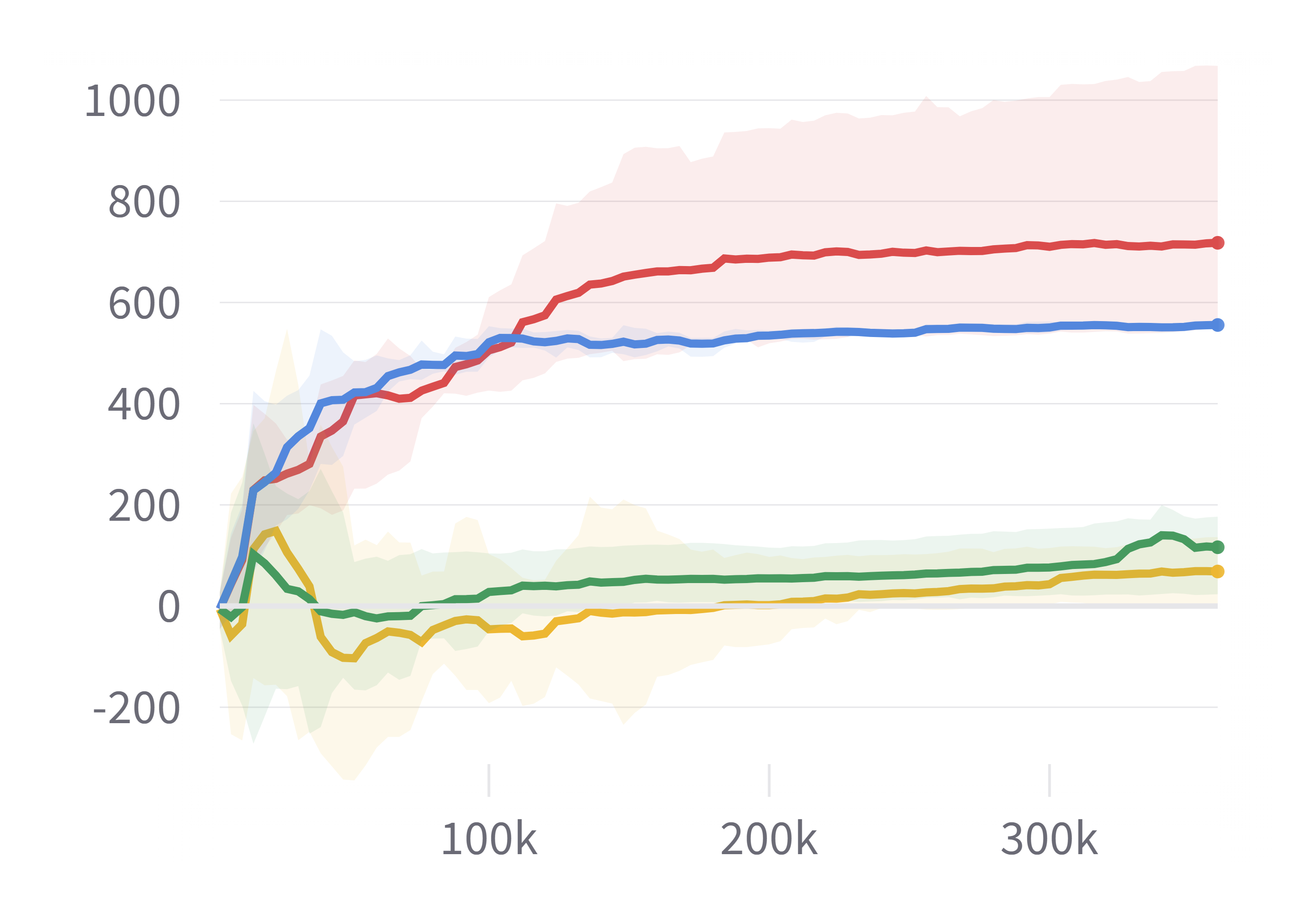}
\end{minipage}
}
\subfigure[Cost]{
\begin{minipage}[t]{0.47\columnwidth}
\centering
\includegraphics[width=1.1\textwidth]{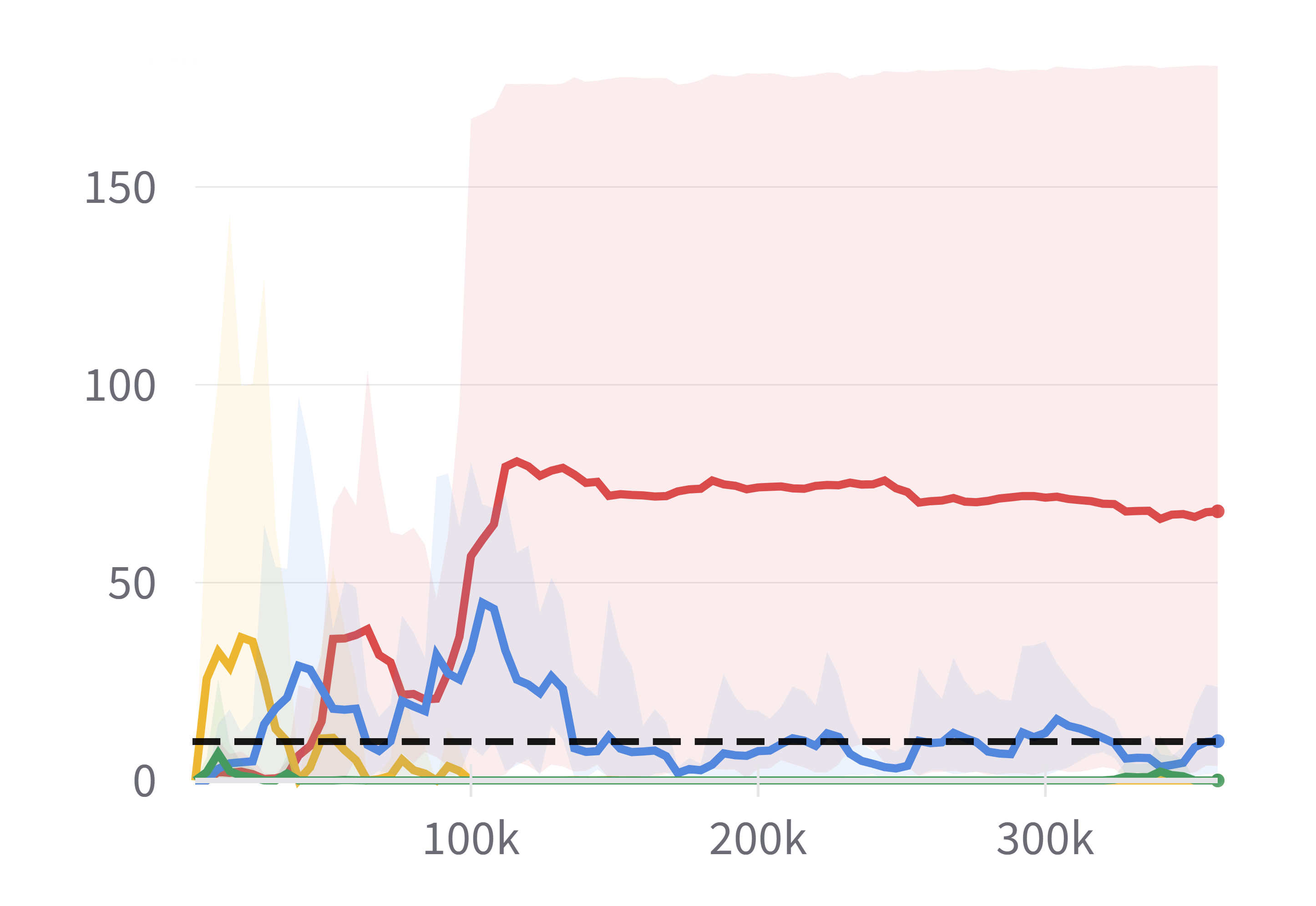}
\end{minipage}
}
\centering
\caption{Learning curves of PPOL over five independent runs with fixed LM values of 1 and 5. The horizontal axis represents time steps. Cost limit $\textbf{d}=10$ (black dashed line) in all experiments. LR = 0.0006 outperforms LR = 0.0003 at LM =1 (red curve is infeasible), while the opposite holds when LM = 5.}
\label{fig_motivation}
\end{figure}

As stated in Section~\ref{section_intro}, despite the impressive capabilities demonstrated by state-of-the-art algorithms like PDO~\cite{chow2017risk} and RCPO~\cite{tessler2018reward} in addressing a wide range of SRL problems, these primal-dual like algorithms still encounter challenges of selecting an appropriate LR for the embedded RL problem during the training process. Indeed, it is natural to posit that employing a constant LR throughout the training process might not be optimal, given that the LM undergoes continuous changes. We illustrate this through a numerical example in  Figure~\ref{fig_motivation}. Figure~\ref{fig_motivation} presents learning curves (both Return and Cost) for the PPOL method over five independent runs using fixed LM values of 1 and 5. We use the \emph{SafetyCarRun-v0} environment from the Bullet-Safety-Gym~\cite{gronauer2022bullet} for this test. Notably, as LM transitions from 1 to 5, the LR needs adjustment from 0.0006 to 0.0003 for optimal results (in terms of return and cost). This means, maintaining a consistent LR across varying LMs might lead to sub-optimal results. Specifically, deploying an optimized LR tailored for an LM value of 5 (green curve) in a scenario meant for LM  value of 1 (red curve) and vice versa leads to compromised performance. 
Therefore, accounting for the dependence between LM and LR is crucial for achieving good performance in SRL.
\subsection{Adaptive Learning Rate}

%




A naive approach to addressing the above challenge is to solve the primal problem in~\eqref{eqn_SRL_problem_lagrangian} for different values of $\lambda$ (LM), each with its optimized LR. However, this approach does not scale when the LM is multi-dimensional (i.e., there are multiple safety constraints) or has a large range. Primal-dual algorithms offer a practical way to find the LM (dual variable) $\lambda^*$ that maximizes the dual function
\begin{align}\label{eqn_dual_function}
 d(\lambda) = \min\limits_{\theta \in \Real^d} \, -J_R(\pi_\theta) + \lambda^T (J_C(\pi_\theta) - {\bf d}).
\end{align}
Consequently the optimal policy (primal variable) $\theta^*$ 
is computed through iterative coordinated updates of the primal and dual variables 
\begin{numcases}{}
\theta_{k+1} = \theta_k - \eta_k \nabla_\theta\mathcal{L}(\theta_k,\lambda_k),  \label{eqn_primal_update}  \\ 
\lambda_{k+1} = \left[\lambda_{k} + \zeta \, g(\theta_{k+1}) \right]_+, \label{eqn_dual_update}
\end{numcases}
where $g(\cdot)$ is the constraint function, i.e., 
\begin{align}\label{eqn_constraint_func}
    g(\theta_{k+1}) \dot{=} J_{C}(\pi_{\theta_{k+1}}) - {\bf d},
\end{align}
and $\eta_k, \zeta$ denote the primal and dual LRs, respectively. Note the dependence of the primal LR $\eta_k$ on iteration $k$ in (\ref{eqn_primal_update}); this is because in our adaptive primal-dual algorithm, described next, $\eta_k$ depends on the LM $\lambda_k$.




We have reached the phase where we articulate the convergence of \eqref{eqn_dual_update}. This is formally established in the following theorem.
\begin{theorem}\label{theorem_dual_convergence}
    Consider the dual function $d(\cdot)$ defined in \eqref{eqn_dual_function}, the constraint function $g(\cdot)$ and cost limit ${\bf d}$ in \eqref{eqn_constraint_func}. Let $\lambda^* \in \argmax_{\lambda \in \mathbb{R}_+^m} d(\lambda) $ and define $D^*= d(\lambda^*) $. 
    Let $\theta_k$ and $\lambda_k$ be the sequences generated by \eqref{eqn_primal_update} and \eqref{eqn_dual_update}. Denote by $\epsilon_k$ the primal error of updating the Lagrangian given $\lambda_k$, i.e., $\epsilon_k = \mathcal{L}(\theta_{k+1}, \lambda_k) -\min_\theta\mathcal{L}(\theta, \lambda_k)$. Define $\lambda_{\text{best}} = \argmax_{\lambda \in \{\lambda_k\}_{k=0}^K }  d(\lambda)$, it holds that
\begin{align}\label{eqn_dual_convergence}
    0 \leq D^* - d\left(\lambda_{\text{best}}\right)
    &\leq  \frac{ \left\| \lambda_{0}- \lambda^* \right\|^2 }{2 \zeta K}  + \frac{\zeta \left( B + \left(1-\gamma\right)\left\|{\bf d}\right\| \right)^2 } {2 \left(1-\gamma\right)^2 } \nonumber \\
    &\quad + \frac{1}{K} \sum_{k=0}^{K-1} \epsilon_{k}.
\end{align}

\end{theorem}
\begin{myproof}
  The proof follows the standard stochastic gradient descent analysis. See the supplementary material for details.
\end{myproof}
Note that the first two terms on the right-hand side of \eqref{eqn_dual_convergence} depend exclusively on the problem and on the dual LR. The last term, however, depends on the primal error $\epsilon_k$. Our goal is to minimize this error through an adaptive selection of the primal LR $\eta_k$. We resort to two different analyses to bound these errors (see Lemma~\ref{lemma_rough_two_bounds}). From these, we derive two adaptive LRs. Before stating Lemma~\ref{lemma_rough_two_bounds}, we require the following assumptions.
\begin{assumption}\label{assumption_Lipschitz}
    Assume the gradients of the objective and cost functions in \eqref{eqn_SRL_problem} are Lipschitz continuous with constants $L_R, L_{C_1}, \cdots, L_{C_m}$. Let $L_C = \left[ L_{C_1}, \cdots, L_{C_m} \right]^T$. This implies that the gradient of Lagrangian~\eqref{eqn_SRL_problem_lagrangian} is Lipschitz on $\theta$
\begin{align}\label{eqn_Lipschitz_lagrangian}
    \left\| \nabla_\theta \mathcal{L}\left(\theta_1, \lambda\right) - \nabla_\theta \mathcal{L}\left(\theta_2, \lambda\right) \right\| \leq L\left(\lambda\right) \left\|\theta_1 - \theta_2\right\|.
\end{align}
where $L\left(\lambda\right) = L_R + \lambda^T L_C$.

\end{assumption}
\begin{assumption}\label{assumption_strong_convex}
 Assume the Lagrangian function $\mathcal{L}\left(\theta, \lambda\right)$ is $\mu$-strongly convex on $\theta$. This is, $\mathcal{L}\left(\theta_{1}, \lambda_k\right) \geq \mathcal{L}\left(\theta_{2}, \lambda_k\right) + \nabla_\theta \mathcal{L}\left(\theta_{2}, \lambda_k\right)^T \\
 \left(\theta_{1} - \theta_{2}\right) + \mu/2 \left\| \theta_{1} - \theta_{2} \right\|^2$.
\end{assumption}


%
%
%
%
%
%
%
%
\begin{lemma}\label{lemma_rough_two_bounds}
    Assume the Lagrangian function \(\mathcal{L}\left(\theta_k, \lambda_k\right)\) is \(L'\)-Lipschitz continuous. Define \(\delta_k = \left\| \theta_{k} - \theta_{k+1}^* \right\|^2\). Then, the following bounds hold for \(\epsilon_k\) given the assumptions that the term inside the square root is non-negative.
%
%
\begin{align}
    \epsilon_k & \leq L' \sqrt{ \frac{2}{\mu}
    \left( L\left(\lambda\right)  \delta_k + \left( L\left(\lambda\right) \left(\eta_k^1\right)^2 - \eta_k^1 \right) \left\|\nabla_\theta \mathcal{L} \left(\theta_k, \lambda_k\right) \right\|^2 \right)}, \label{eqn_invlin_tight_bound} \\
    \epsilon_k & \leq L' \sqrt{ \left(1+\left(\eta_k^2\right)^2 L\left(\lambda\right)^2- \eta_k^2 \mu\right) \delta_k}. \label{eqn_invqua_tight_bound}
\end{align}
\end{lemma}
\begin{myproof}
We start the proof by employing the \(L'\)-Lipschitz continuity of the Lagrangian to bound \(\epsilon_k\)
\begin{equation}\label{eqn_epsilon_diff_theta}
    \epsilon_k=\mathcal{L}\left(\theta_{k+1},\lambda_{k}\right) -\mathcal{L}\left(\theta_{k+1}^*,\lambda_{k}\right)\leq L' \left\|\theta_{k+1}-\theta_{k+1}^*\right\|.
\end{equation}
Hence, to bound \(\epsilon_k\), one can instead investigate \(\left\|\theta_{k+1}-\theta_{k+1}^*\right\|\).

We first focus on \eqref{eqn_invlin_tight_bound}. It follows from the strong convexity of the Lagrangian with respect to \(\theta\) that
\begin{align}
    \mathcal{L} \left(\theta_{k+1}, \lambda_k\right) &\geq \mathcal{L} \left(\theta_{k+1}^*, \lambda_k\right) + \nabla_\theta \mathcal{L} \left(\theta_{k+1}^*, \lambda_k\right)^T \left(\theta_{k+1} - \theta_{k+1}^*\right) \nonumber \\
    &\quad + \frac{\mu}{2} \left\| \theta_{k+1} - \theta_{k+1}^* \right\|^2.
\end{align}
Since \(\nabla_\theta \mathcal{L} \left(\theta_{k+1}^*, \lambda_k\right) = 0\) the previous inequality reduces to
\begin{align}\label{eqn_auxiliary_equation}
    \left\| \theta_{k+1} - \theta_{k+1}^* \right\|^2 &\leq \frac{2}{\mu}
    \left( \mathcal{L} \left(\theta_{k+1}, \lambda_k\right) - \mathcal{L} \left(\theta_{k+1}^*, \lambda_k\right) \right).
\end{align}

We apply the Taylor expansion on both terms on the right-hand side of the previous expression around $\theta_k$
\begin{equation}\label{eqn_taylor1}
    \mathcal{L} (\theta_{k+1}, \lambda_k) = \mathcal{L} (\theta_{k}, \lambda_k) + \nabla_\theta \mathcal{L} (\theta_c, \lambda_k)^T \left(\theta_{k+1}-\theta_k\right),
\end{equation}
where $\theta_c = \xi \theta_k + (1-\xi) \theta_{k+1}, \, \xi^\in[0,1]$, and 
\begin{equation}\label{eqn_taylor2}
    \mathcal{L} (\theta_{k+1}^*, \lambda_k) = \mathcal{L} (\theta_{k}, \lambda_k) + \nabla_\theta \mathcal{L} (\theta_c^\prime, \lambda_k)^T \left(\theta_{k+1}^* - \theta_k\right),
\end{equation}
where $\theta_c^\prime = \xi^\prime \theta_k + (1-\xi^\prime) \theta_{k+1}^*, \, \xi^\prime\in[0,1]$.
Substituting \eqref{eqn_taylor1} and \eqref{eqn_taylor2} into \eqref{eqn_auxiliary_equation} reduces to
\begin{align}\label{eqn_long_inequality}
    \left\| \theta_{k+1} - \theta_{k+1}^*\right\|^2 &\leq \frac{2}{\mu}\left(\nabla_\theta \mathcal{L} (\theta_c', \lambda_k)^T (\theta_k -\theta_{k+1}^*) \right. \\ \nonumber
    &\quad + \left.\nabla_\theta \mathcal{L} (\theta_c, \lambda_k)^T (\theta_{k+1} - \theta_k)\right).
\end{align}


By virtue of \(\nabla_\theta \mathcal{L} \left(\theta_{k+1}^*, \lambda_k\right) = 0\) \eqref{eqn_long_inequality} reduces to
\begin{align}
    \left\| \theta_{k+1} - \theta_{k+1}^* \right\|^2 &\leq \frac{2}{\mu}
    \left( \left(\nabla_\theta \mathcal{L} \left(\theta_c', \lambda_k\right)-\nabla_\theta \mathcal{L} \left(\theta_{k+1}^*, \lambda_k\right)\right)^T \right. \nonumber \\
    & \quad \left. \left(\theta_k -\theta_{k+1}^*\right) + \nabla_\theta \mathcal{L} \left(\theta_c, \lambda_k\right)^T \left(\theta_{k+1} - \theta_k  \right) \right)  .
\end{align}
where the last inequality follows from the Lipschitz continuity of the gradient of Lagrangian (see Assumption \ref{assumption_Lipschitz}).

By adding and subtracting \(\nabla_\theta \mathcal{L} \left(\theta_k, \lambda_k\right)^T \left(\theta_{k+1} - \theta_k\right)\) we obtain
\begin{align}
    \left\| \theta_{k+1} - \theta_{k+1}^* \right\|^2 &\leq \frac{2}{\mu}
    \left( L(\lambda) \left\|\theta_k -\theta_{k+1}^*\right\|^2 + \right. \nonumber \\
    & \left(\nabla_\theta \mathcal{L} \left(\theta_c, \lambda_k\right) - \nabla_\theta \mathcal{L} \left(\theta_k, \lambda_k\right)\right)^T \left(\theta_{k+1} - \theta_k\right) \nonumber \\
    & \left. +\nabla_\theta \mathcal{L} \left(\theta_k, \lambda_k\right)^T \left(\theta_{k+1} - \theta_k\right) \right).
\end{align}
Employing Lipschitz continuity of the gradient of Lagrangian yields
\begin{align}
    \left\| \theta_{k+1} - \theta_{k+1}^* \right\|^2 &\leq \frac{2}{\mu}
    \left( L(\lambda) \left\|\theta_k -\theta_{k+1}^*\right\|^2 + L(\lambda) \left\|\theta_{k+1} - \theta_k\right\|^2 \right. \nonumber \\
    & \left. +\nabla_\theta \mathcal{L} \left(\theta_k, \lambda_k\right)^T \left(\theta_{k+1} - \theta_k\right) \right).
\end{align}
By definition of \(\delta_k\) the previous inequality can be rewritten as
\begin{align}
    \left\| \theta_{k+1} - \theta_{k+1}^* \right\|^2 & \leq \frac{2}{\mu}
    \left( L(\lambda)  \delta_k + L(\lambda) \left(\eta_k^1\right)^2 \left\| \nabla_\theta \mathcal{L} \left(\theta_k, \lambda_k\right) \right\|^2 \right.\nonumber \\
    & \quad - \left. \eta_k^1 \left\|\nabla_\theta \mathcal{L} \left(\theta_k, \lambda_k\right) \right\|^2 \right).
\end{align}

Taking the square root on the previous inequality and combining with \eqref{eqn_epsilon_diff_theta} completes the proof of \eqref{eqn_invlin_tight_bound}. 

We then turn our attention to proving \eqref{eqn_invqua_tight_bound}. 
%
Note that
\begin{align}
    \left\|  \theta_{k+1} - \theta_{k+1}^*  \right\|^2 &= \left\|  \theta_{k} - \eta_k^2 \nabla_\theta \mathcal{L} \left(\theta_k, \lambda_k\right) - \theta_{k+1}^*  \right\|^2 \nonumber \\
    &= \left\|  \theta_{k} - \theta_{k+1}^* \right\|^2 + \left(\eta_k^2\right)^2 \left\| \nabla_\theta \mathcal{L} \left(\theta_k, \lambda_k\right) \right\|^2 \nonumber \\
    &\quad - 2 \eta_k^2 \nabla_\theta \mathcal{L} \left(\theta_k, \lambda_k\right)^T \left(\theta_k - \theta_{k+1}^*\right).
\end{align}
By \(\nabla_\theta \mathcal{L} \left(\theta_{k+1}^*, \lambda_k\right) = 0\) the previous equation is equivalent to
\begin{align}
    \left\|  \theta_{k+1} - \theta_{k+1}^*  \right\|^2
    &= \left\|  \theta_{k} - \theta_{k+1}^* \right\|^2  \nonumber \\
    &+ \left(\eta_k^2\right)^2 \left\| \nabla_\theta \mathcal{L} \left(\theta_k, \lambda_k\right) - \nabla_\theta \mathcal{L} \left(\theta_{k+1}^*, \lambda_k\right) \right\|^2 \nonumber \\
    &- 2 \eta_k^2 \nabla_\theta \mathcal{L} \left(\theta_k, \lambda_k\right)^T \left(\theta_k - \theta_{k+1}^*\right).
\end{align}
Assumption~\ref{assumption_Lipschitz} indicates that the gradient of Lagrangian function is \(L(\lambda)\)-Lipschitz continuous, and we thus obtain
\begin{align}\label{eqn_blabla1}
    \left\|  \theta_{k+1} - \theta_{k+1}^*  \right\|^2
    &\leq \left\|  \theta_{k} - \theta_{k+1}^* \right\|^2 \nonumber \\
    &\quad + \left(\eta_k^2\right)^2 L(\lambda)^2 \left\| \theta_k - \theta_{k+1}^* \right\|^2 \nonumber \\
    &\quad - 2 \eta_k^2 \nabla_\theta \mathcal{L} \left(\theta_k, \lambda_k\right)^T \left(\theta_k - \theta_{k+1}^*\right).
\end{align}

In addition, the strong convexity of the Lagrangian with respect to $\theta$ shows that 
\begin{align}\label{eqn_blabla2}
    \nabla_\theta \mathcal{L} \left(\theta_k, \lambda_k\right)^T \left(\theta_{k+1}^* - \theta_k\right)  &\leq   \mathcal{L} \left(\theta_{k+1}^*, \lambda_k\right)  - \mathcal{L} \left(\theta_k, \lambda_k\right) \nonumber \\
    & \quad - \frac{\mu}{2} \left\| \theta_{k+1}^* -\theta_k\right\|^2 \nonumber  \\
    &\leq - \frac{\mu}{2} \left\| \theta_{k+1}^* -\theta_k\right\|^2,
\end{align}
where the last inequality follows from the fact that \(\theta_{k+1}^*\) is the minimizer of the Lagrangian. Subsequently, combining \eqref{eqn_blabla1} and \eqref{eqn_blabla2} yields
\begin{align}
    \left\|  \theta_{k+1} - \theta_{k+1}^*  \right\|^2
    &\leq \left\|  \theta_{k} - \theta_{k+1}^* \right\|^2 - \eta_k^2 \mu \left\|\theta_k - \theta_{k+1}^*\right\|^2\nonumber \\
    & \quad + \left(\eta_k^2\right)^2 L(\lambda)^2 \left\| \theta_k - \theta_{k+1}^* \right\|^2 \nonumber \\
    &=\left(1+\left(\eta_k^2\right)^2 L(\lambda)^2- \eta_k^2 \mu\right) \left\|\theta_k - \theta_{k+1}^*\right\|^2.
\end{align}
Taking the square root and combining with \eqref{eqn_epsilon_diff_theta} completes the proof of the result.
\end{myproof}

By virtue of Assumptions~\ref{assumption_Lipschitz}, ~\ref{assumption_strong_convex} and Lemma~\ref{lemma_rough_two_bounds}, we are able to derive two optimal LRs that minimize the two bounds in \eqref{eqn_invlin_tight_bound} and \eqref{eqn_invqua_tight_bound}. We formalize this claim in the following theorem.

%

\begin{theorem}\label{theorem_invlinqua_bound}
    Let Assumption~\ref{assumption_Lipschitz} and Assumption~\ref{assumption_strong_convex} hold. Then, the learning rates \(\eta_k^1 = \frac{1}{2 L\left(\lambda\right)}\) and \(\eta_k^2 =  \frac{\mu}{2L\left(\lambda\right)^2}\) minimize the bounds in \eqref{eqn_invlin_tight_bound} and \eqref{eqn_invqua_tight_bound}, respectively. Denote by \(\epsilon_{k}^1\) and \(\epsilon_{k}^2\) the bounds on the primal error using \(\eta_k^1\) and \(\eta_k^2\). Consider the \(L'\)-Lipschitz continuity of the Lagrangian \(\mathcal{L}\left(\theta_k, \lambda_k\right)\) and \(\delta_k\) in Lemma~\ref{lemma_rough_two_bounds}. Then, it holds that
\begin{align}
    \epsilon_{k}^1 &\leq L' \sqrt{ \frac{2 \delta_k}{\mu}
    \left( L\left(\lambda\right) - \frac{ \mu^2}{16 L\left(\lambda\right)} \right)}, \label{eqn_bound_inverse_linear} \\
    \epsilon_{k}^2 &\leq L' \sqrt{ \delta_k \left( 1 - \frac{\mu^2}{4 L\left(\lambda\right)^2} \right)}. \label{eqn_bound_inverse_quadratic}
\end{align}
\end{theorem}

\begin{myproof}
We start by establishing two LRs $\eta_k^1$ and $\eta_k^2$. As observed in \eqref{eqn_invlin_tight_bound}, the right-hand side is convex with respect to $\eta_k^1$ since $L', L(\lambda), \delta_k$ are independent of $\eta_k^1$. To minimize it one can compute the derivative and make it zero. Then the minimizer of the right-hand side of \eqref{eqn_invlin_tight_bound} is given by
%
%
\begin{align}\label{eqn_invlin_expression}
    \eta_k^1 = \frac{1}{2 L(\lambda)}.
\end{align}


Likewise, the right-hand side of \eqref{eqn_invqua_tight_bound} is convex with respect to $\eta_k^2$, and the optimal LR (minimizer) takes the form of
\begin{align}\label{eqn_invqua_expression}
    \eta_k^2 =  \frac{\mu}{2L(\lambda)^2}.
\end{align}

Having established \( \eta_k^1 \) and \( \eta_k^2 \), we are now in the stage of proving \( \epsilon_k^1 \) and \( \epsilon_k^2 \). For \( \epsilon_k^1 \), substituting \( \eta_k^1 \) into \eqref{eqn_invlin_tight_bound} yields the tightest upper bound, i.e.,
\begin{align}\label{eqn_invlin_tight_bound2}
    \epsilon_k^1 \leq L' \sqrt{ \frac{2}{\mu}
    \left( L(\lambda)  \delta_k - \frac{\left\| \nabla_\theta \mathcal{L} (\theta_k, \lambda_k) \right\|^2}{4 L(\lambda)} \right)}.
\end{align}


Strong convexity indicates that
\begin{align}
    -\nabla_\theta \mathcal{L} \left(\theta_k, \lambda_k\right)^T  \left(\theta_{k+1}^* - \theta_k\right) \geq \frac{\mu}{2} \left\| \theta_{k+1}^* - \theta_k \right\|^2.
\end{align}
Squaring both sides of the previous inequality, using the Cauchy-Schwartz inequality and the definition of \(\delta_k\) it follows that
\begin{align}
    \left\| \nabla_\theta \mathcal{L} \left(\theta_k, \lambda_k\right) \right\|^2 \geq \frac{\mu^2}{4} \left\| \theta_{k+1}^* - \theta_k \right\|^2 = \frac{\mu^2}{4} \delta_k.
\end{align}

Combining the previous inequality with \eqref{eqn_invlin_tight_bound2} yields
%
\begin{align}
    \epsilon_{k}^1 \! \leq \! L' \sqrt{ \frac{2 \delta_k}{\mu}
    \left( L(\lambda) - \frac{ \mu^2}{16 L(\lambda)} \right) }.
\end{align}

Our focus now shifts towards the derivation of $\epsilon_k^2$. Substituting $\eta_k^2$ into \eqref{eqn_invqua_tight_bound} yields the tightest upper bound on $\epsilon_k^2$
%
\begin{align}
    \epsilon_{k}^2  \leq  L'  \sqrt{ \delta_k \left( 1 - \frac{\mu^2}{4 L(\lambda)^2} \right) }.
\end{align}
These complete the proof of Theorem~\ref{theorem_invlinqua_bound}. 
\end{myproof}

%
%
%
\begin{algorithm}[tb]
\caption{Adaptive Primal-Dual (APD)}
\label{alg_apd}
\begin{algorithmic}[1] 
\STATE \textbf{Input}: \( \theta_0,\lambda_0, \zeta, L_R, L_C, \mu \) (optional)
\FOR{\( k = 0, 1, \cdots, \)}
    \STATE Choose primal LR from \eqref{eqn_invlin_expression} or \eqref{eqn_invqua_expression}
    \STATE Update primal variable (policy parameter) as in \eqref{eqn_primal_update}
    \STATE Update dual variable (LM) as in \eqref{eqn_dual_update}
\ENDFOR
\end{algorithmic}
\end{algorithm}
\begin{algorithm}[tb]
\caption{Practical Adaptive Primal-Dual (PAPD)}
\label{alg_papd}
\begin{algorithmic}[1] 
\STATE \textbf{Input}: \( \theta_0, H_1 (H_1'), H_2 (H_2'), K_P ,K_I ,K_D \)
\FOR{\( k = 0, 1, \cdots, \)}
    \STATE Choose primal LR from \eqref{eqn_two_lr_prac}
    \STATE Update primal variable (policy parameter) as in \eqref{eqn_primal_update}
    \STATE Update dual variable (LM) as in \eqref{eqn_pid_lagrangian}
\ENDFOR
\end{algorithmic}
\end{algorithm}
Notice that both LRs proposed by Theorem~\ref{theorem_invlinqua_bound} demonstrate an inverse relationship with respect to the LM, which is also empirically validated by our preliminary observations in Figure~\ref{fig_motivation}. Moreover, $\eta_{k}^1$ has an inverse-linear dependence on LM while $\eta_{k}^2$ has an inverse-quadratic dependence on LM. We therefore term them \emph{InvLin} and \emph{InvQua}, respectively. It is important to point out that we assume that the Lagrangian is strongly convex. This is generally not the case for RL problems. However, one can assume local convexity around a local minimum. We chose this stronger assumption for simplicity in the exposition.

Having established \emph{InvLin} and \emph{InvQua} as well as corresponding bounds $\epsilon_k^1$ and $\epsilon_k^2$, we are able to propose an adaptive primal-dual (APD) algorithm, which is summarized under Algorithm~\ref{alg_apd}. Notice that Theorem~\ref{theorem_dual_convergence} indicating the (approximate) convergence of the LM also holds for the APD algorithm. Furthermore, with $\epsilon_k^1$ and $\epsilon_k^2$, we can further establish the guarantee of proximity to the primal optimum $J_R(\pi_{\theta^*})$. The following theorem addresses this aspect explicitly.
\begin{theorem}\label{theorem_primal_bound}
    Consider \( \theta^* \) and \( J_R(\pi_{\theta}) \) in \eqref{eqn_SRL_problem}. Let the hypotheses of Theorem~\ref{theorem_dual_convergence} hold, and consider the sequence of the LM generated by Algorithm~\ref{alg_apd}. Then, it holds that
    \begin{align}\label{eqn_primal_convergence}
    \liminf_{K \to \infty} \frac{1}{K} \sum_{k=0}^{K-1} J_R(\pi_{\theta_{k+1}}) &\geq J_R(\pi_{\theta^*}) - \frac{1}{K} \left( \sum_{k=0}^{K-1} \epsilon_{k} \right) \nonumber \\
    &\quad  - \frac{\zeta \left( B + (1-\gamma) \left\|{\bf d}\right\| \right)^2 } {2 (1-\gamma)^2 }.
    \end{align}
\end{theorem}

\begin{myproof}
   See the supplemental material.
\end{myproof}
Notice that $\epsilon_{k}$ in \eqref{eqn_primal_convergence} can be bounded by \eqref{eqn_bound_inverse_linear} or \eqref{eqn_bound_inverse_quadratic}, depending on \emph{InvLin} or \emph{InvQua} is selected. Theorem~\ref{theorem_primal_bound} demonstrates that the limit inferior of the average of the sequence derived by Algorithm~\ref{alg_apd} approximates well the value of $J_R(\pi_{\theta^*})$. In principle, the limit superior has the potential to be significantly larger than $J_R(\pi_{\theta^*})$, which would lead to constraint violations. In the following result, we prove that this is not the case, i.e., the sequence generated by Algorithm~\ref{alg_apd} is feasible on average.


\begin{theorem}
\label{theorem_feasibility}
Let hypotheses of Theorem~\ref{theorem_dual_convergence} hold. It holds that
\begin{align}
    \limsup_{K \to \infty}  \frac{1}{K} \sum_{k=0}^{K-1} J_{C_i}(\pi_{\theta_k}) \leq  d_i, \, i=1, 2, \cdots, m.
\end{align}
\end{theorem}
\begin{myproof}
   See the supplemental material.
\end{myproof}

Despite the theoretical guarantees on APD (Algorithm~\ref{alg_apd}) regarding the convergence, optimality, and feasibility, estimating the Lipschitz constant $L(\lambda)$ and the strongly convex constant $\mu$ in the LRs~\eqref{eqn_invlin_expression} and \eqref{eqn_invqua_expression} is, in general, computationally expensive. We thus consider a practical version of adaptive LRs. Under the single constraint, the adaptive LRs in \eqref{eqn_invlin_expression} and \eqref{eqn_invqua_expression} can be written as
\begin{align}\label{eqn_two_lr_prac}
    \eta_k^1 = H_1 / (\lambda_k + H_2), \quad \eta_k^2 = H_1' / (\lambda_k + H_2')^2,
\end{align}
where $H_1, H_2, H_1', H_2'$ are hyper-parameters.

Note that the update on the policy parameter is updated by running a step of any RL algorithm with the adaptive LRs in \eqref{eqn_two_lr_prac}. In particular, we consider state-of-the-art algorithms such as PPOL and DDPGL. However, the dual variables frequently exhibit tendencies to overshoot and oscillate, thereby hindering performance. To address these challenges, we adopt the Proportional Integral Derivative (PID) Lagrangian strategies as introduced by \cite{stooke2020responsive}. Drawing inspiration from the feedback control, the LM is updated as
\begin{align}\label{eqn_pid_lagrangian}
\lambda_k  = \left( K_P (J_C^k-d)+K_I I^k+K_D (J_C^k-J_C^{k-1}) \right)_{+},
\end{align}
where $I^k$ is recursively defined as $I^k = (I^{k-1}+J^k_C-d)_+$ and subscript \( + \) indicates projection onto the non-negative orthant. The three hyper-parameters \( K_P \), \( K_I \), and \( K_D \) represent the proportional, integral, and derivative gains. In particular, selecting $K_P=K_D=0$ the update reduces to gradient descent on the dual domain.

With the practical adaptive LRs (\emph{InvLin} and \emph{InvQua}) defined as in \eqref{eqn_two_lr_prac}, as well as the dual update rule \eqref{eqn_pid_lagrangian} using PID-Lagrangian, the practical version of APD (PAPD) algorithm, which we implement and evaluate in the subsequent section, is described in Algorithm~\ref{alg_papd}.

\begin{figure*}
\centering
\subfigure{
\begin{minipage}[t]{0.15\linewidth}
\centering
\includegraphics[width=0.7\textwidth]{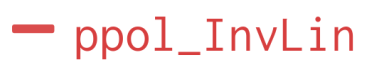}
\end{minipage}
}
\subfigure{
\begin{minipage}[t]{0.15\linewidth}
\centering
\includegraphics[width=0.7\textwidth]{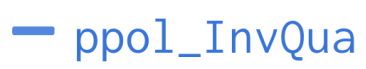}
\end{minipage}
}
\subfigure{
\begin{minipage}[t]{0.15\linewidth}
\centering
\includegraphics[width=0.7\textwidth]{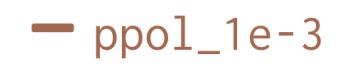}
\end{minipage}
}
\subfigure{
\begin{minipage}[t]{0.15\linewidth}
\centering
\includegraphics[width=0.7\textwidth]{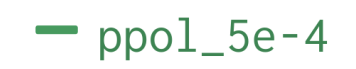}
\end{minipage}
}
\subfigure{
\begin{minipage}[t]{0.15\linewidth}
\centering
\includegraphics[width=0.7\textwidth]{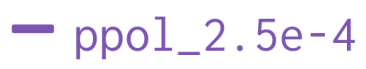}
\end{minipage}
}
\subfigure{
\begin{minipage}[t]{0.15\linewidth}
\centering
\includegraphics[width=0.7\textwidth]{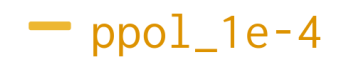}
\end{minipage}
}
\subfigure{
\begin{minipage}[t]{0.23\linewidth}
\centering
\includegraphics[width=1.1\textwidth]{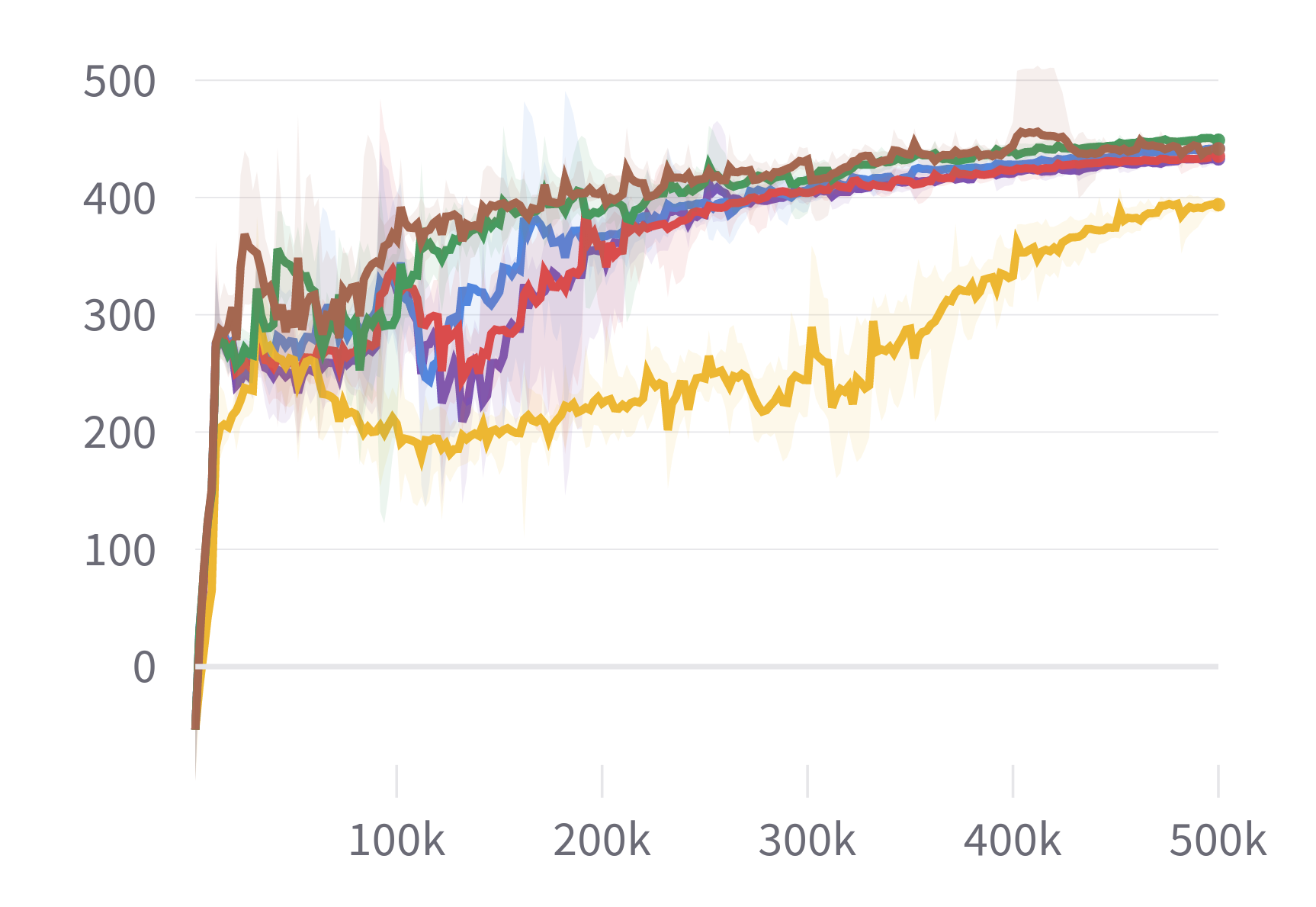}
\caption*{\small{Return-\emph{SafetyBallRun-v0}}}
\end{minipage}
}
\subfigure{
\begin{minipage}[t]{0.23\linewidth}
\centering
\includegraphics[width=1.1\textwidth]{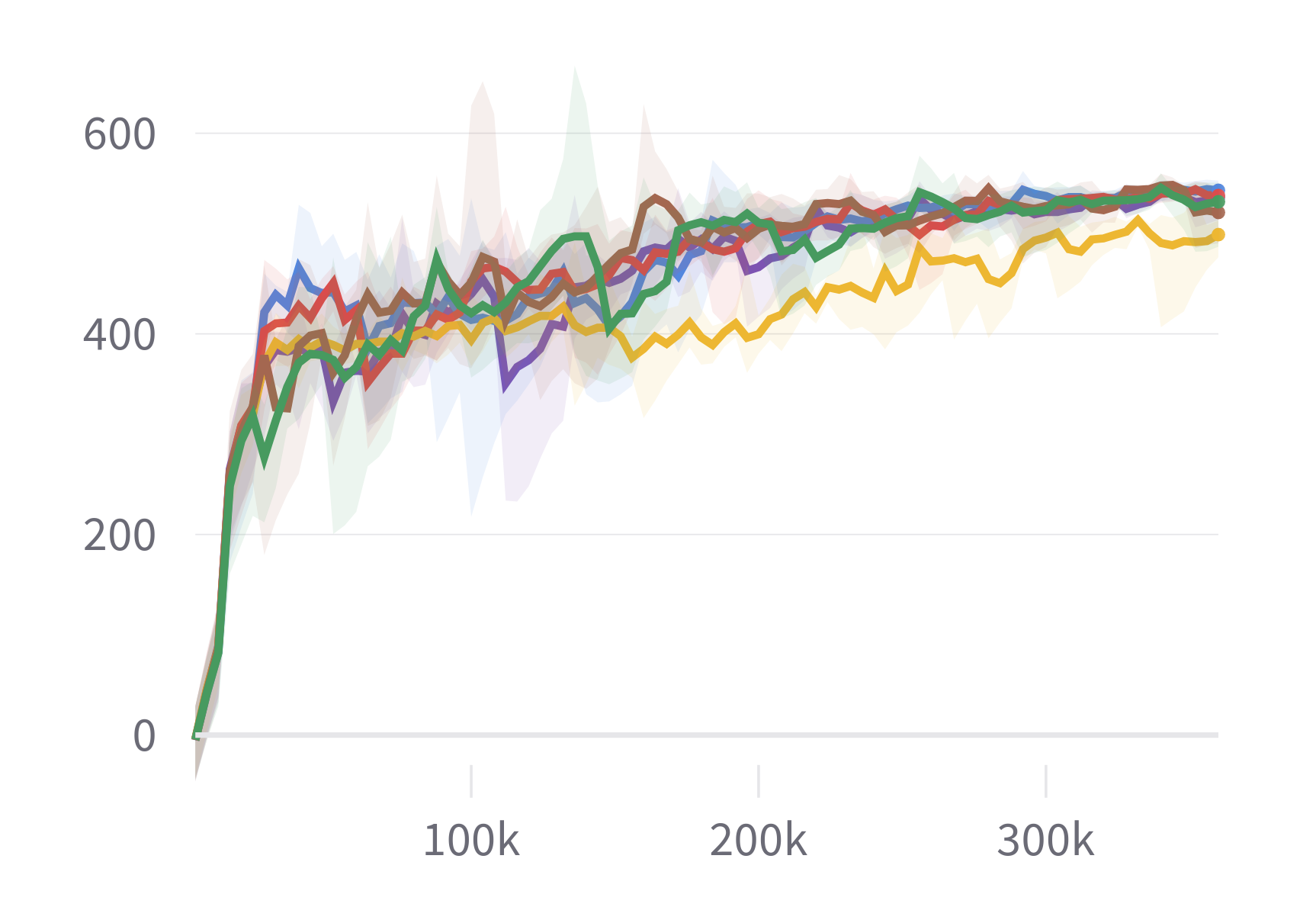}
\caption*{\small{Return-\emph{SafetyCarRun-v0}}}
\end{minipage}
}
\subfigure{
\begin{minipage}[t]{0.23\linewidth}
\centering
\includegraphics[width=1.1\textwidth]{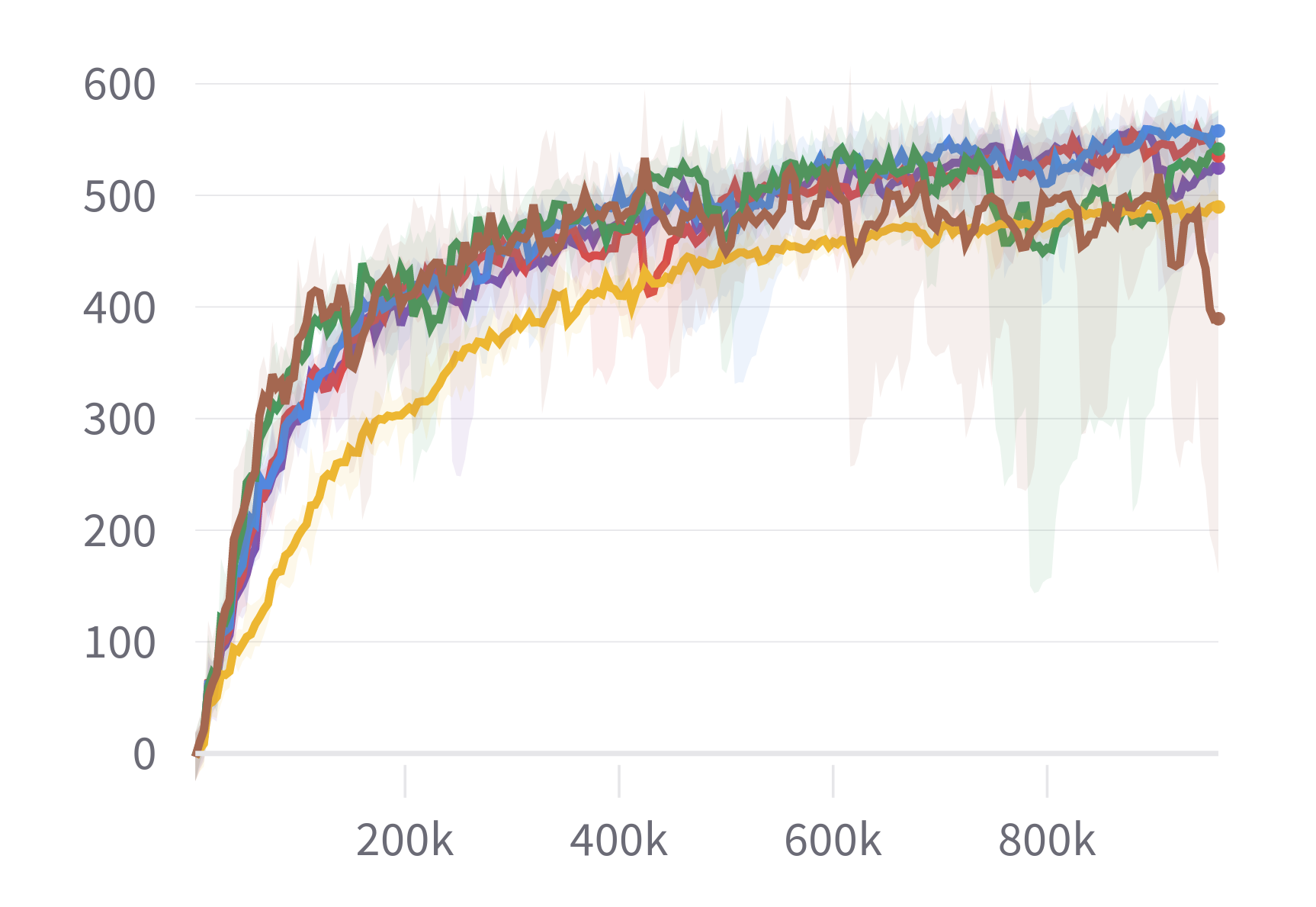}
\caption*{\small{Return-\emph{SafetyBallCircle-v0}}}
\end{minipage}
}
\subfigure{
\begin{minipage}[t]{0.23\linewidth}
\centering
\includegraphics[width=1.1\textwidth]{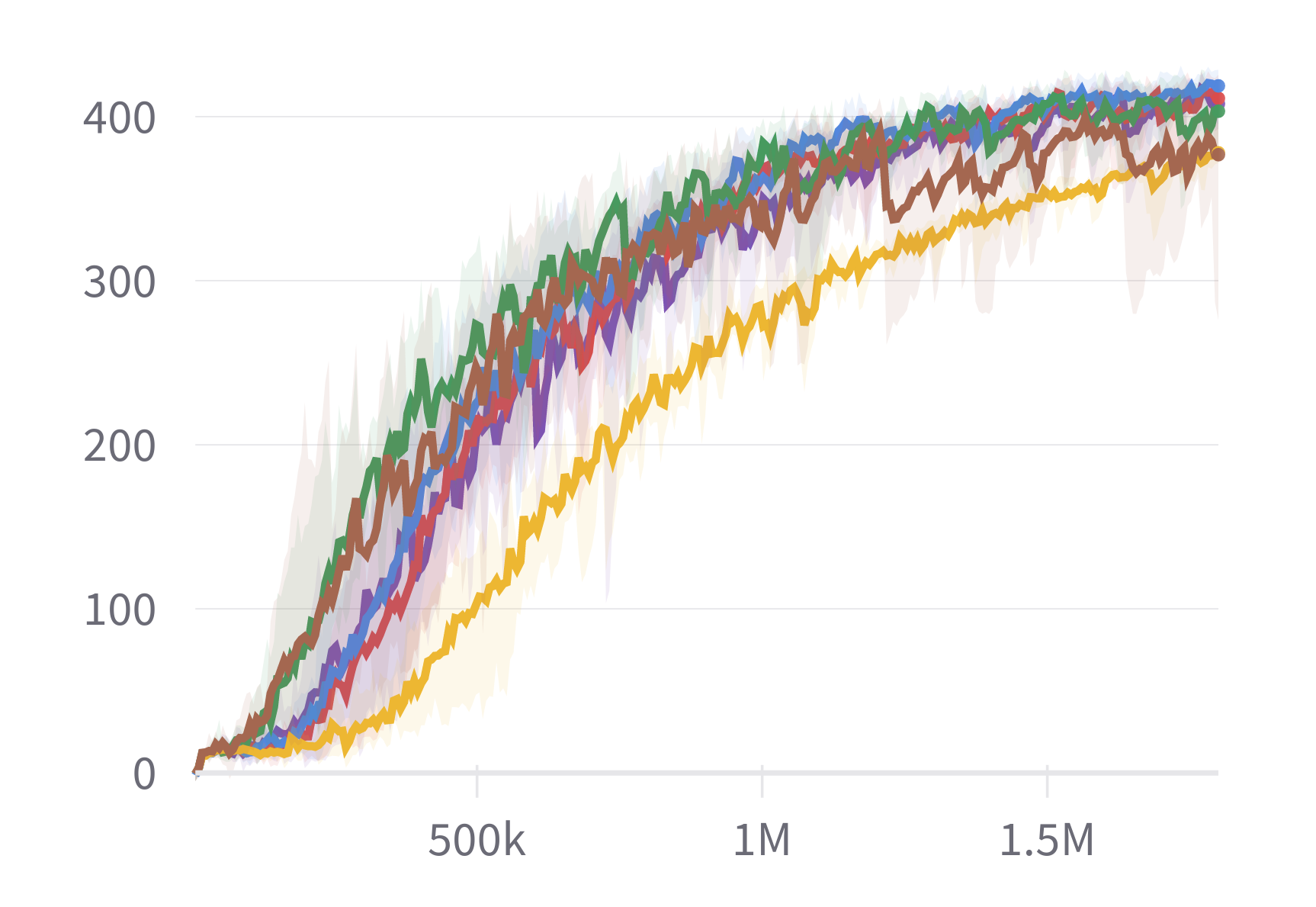}
\caption*{\small{Return-\emph{SafetyCarCircle-v0}}}
\end{minipage}
}
\subfigure{
\begin{minipage}[t]{0.23\linewidth}
\centering
\includegraphics[width=1.1\textwidth]{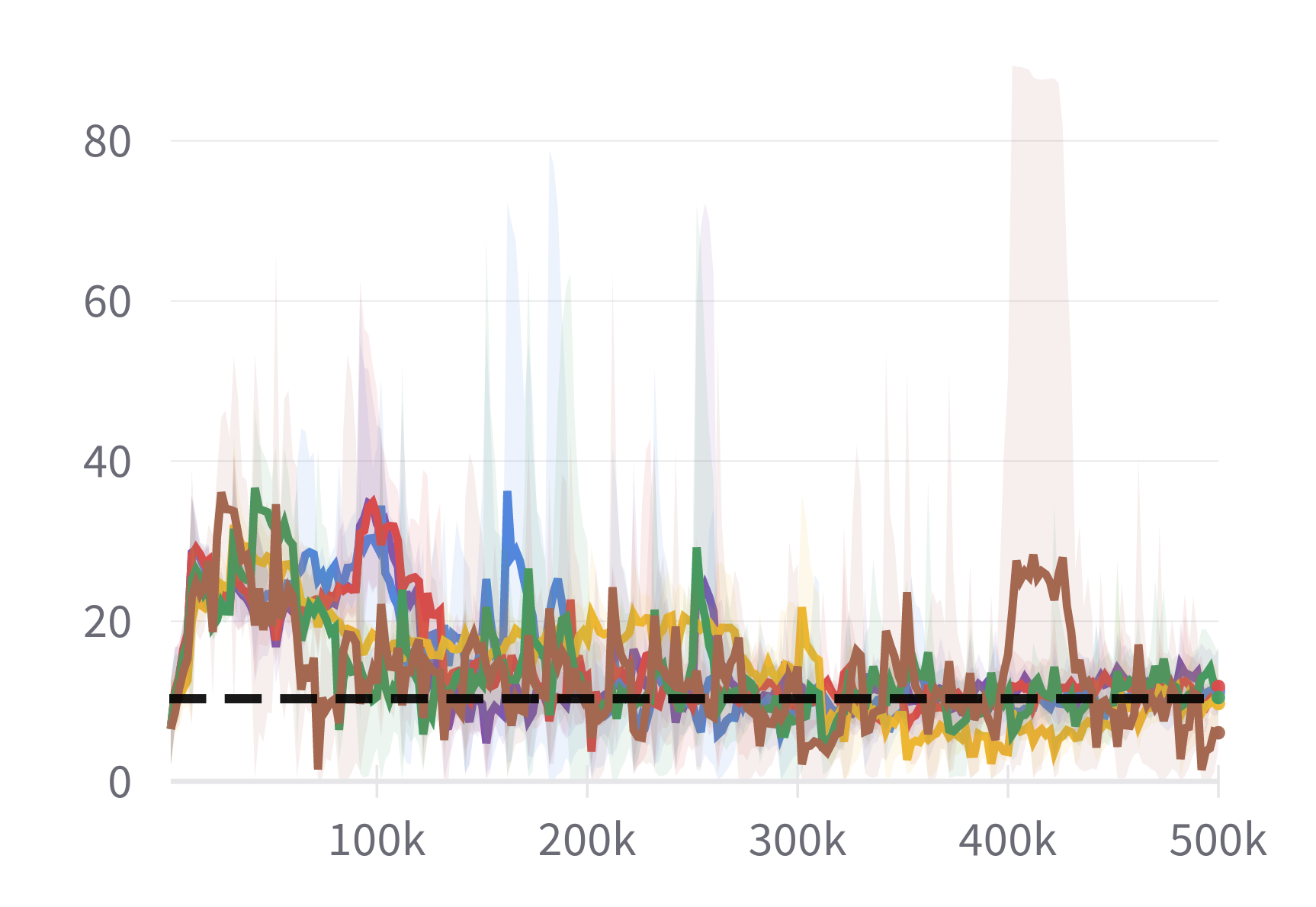}
\caption*{\small{Cost-\emph{SafetyBallRun-v0}}}
\end{minipage}
}
\subfigure{
\begin{minipage}[t]{0.23\linewidth}
\centering
\includegraphics[width=1.1\textwidth]{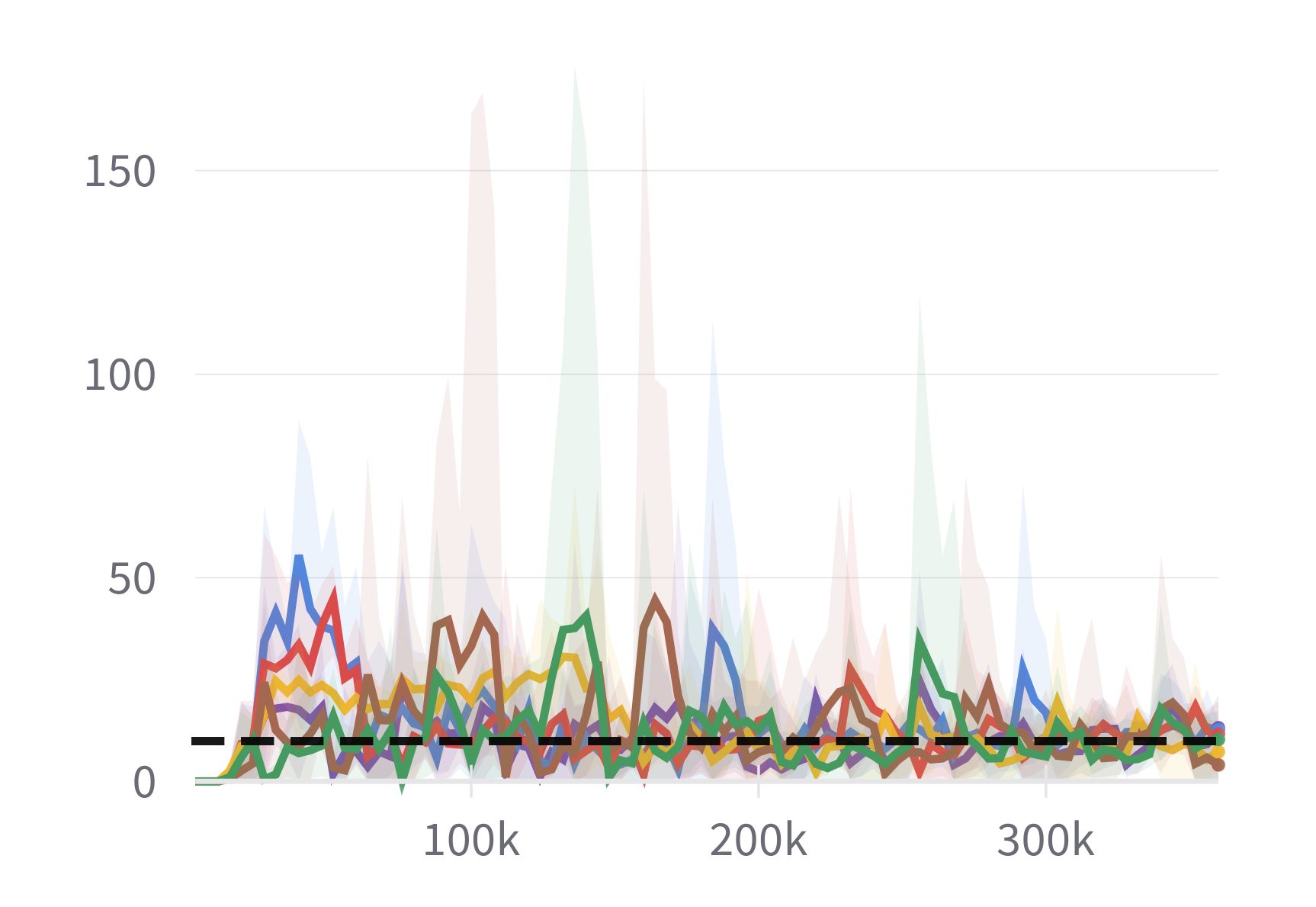}
\caption*{\small{Cost-\emph{SafetyCarRun-v0}}}
\end{minipage}
}
\subfigure{
\begin{minipage}[t]{0.23\linewidth}
\centering
\includegraphics[width=1.1\textwidth]{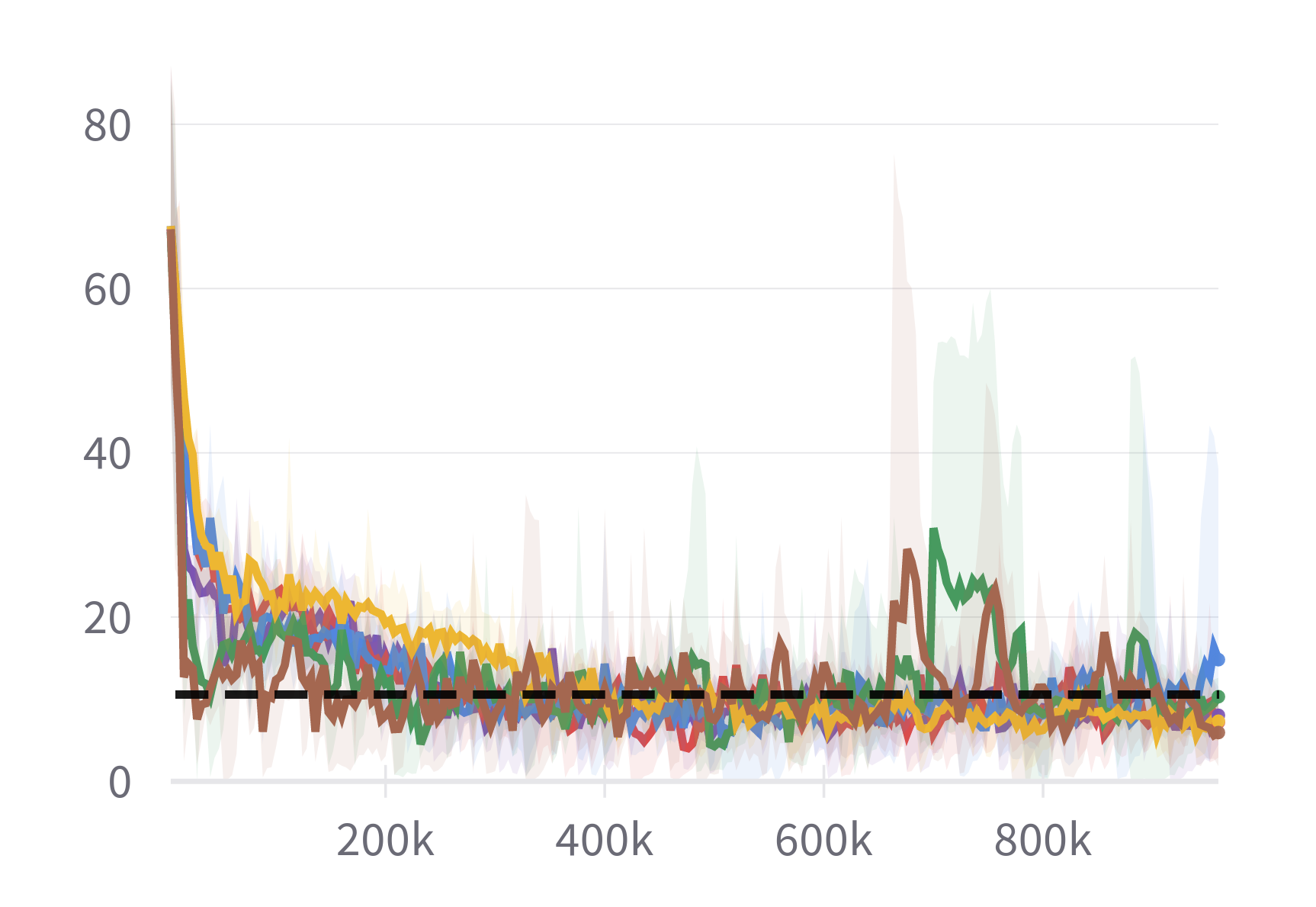}
\caption*{\small{Cost-\emph{SafetyBallCircle-v0}}}
\end{minipage}
}
\subfigure{
\begin{minipage}[t]{0.23\linewidth}
\centering
\includegraphics[width=1.1\textwidth]{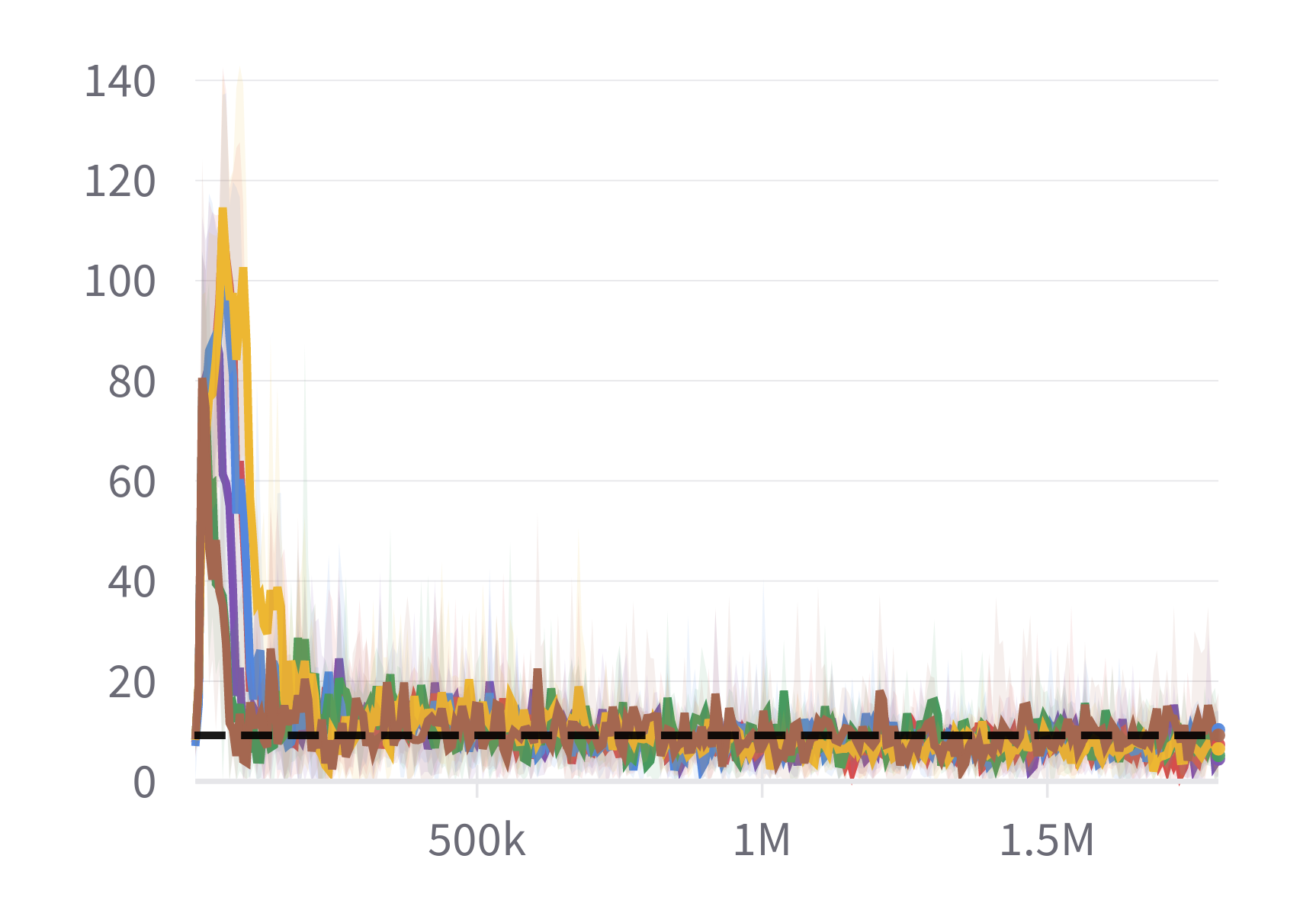}
\caption*{\small{Cost-\emph{SafetyCarCircle-v0}}}
\end{minipage}
}
\subfigure{
\begin{minipage}[t]{0.23\linewidth}
\centering
\includegraphics[width=1.1\textwidth]{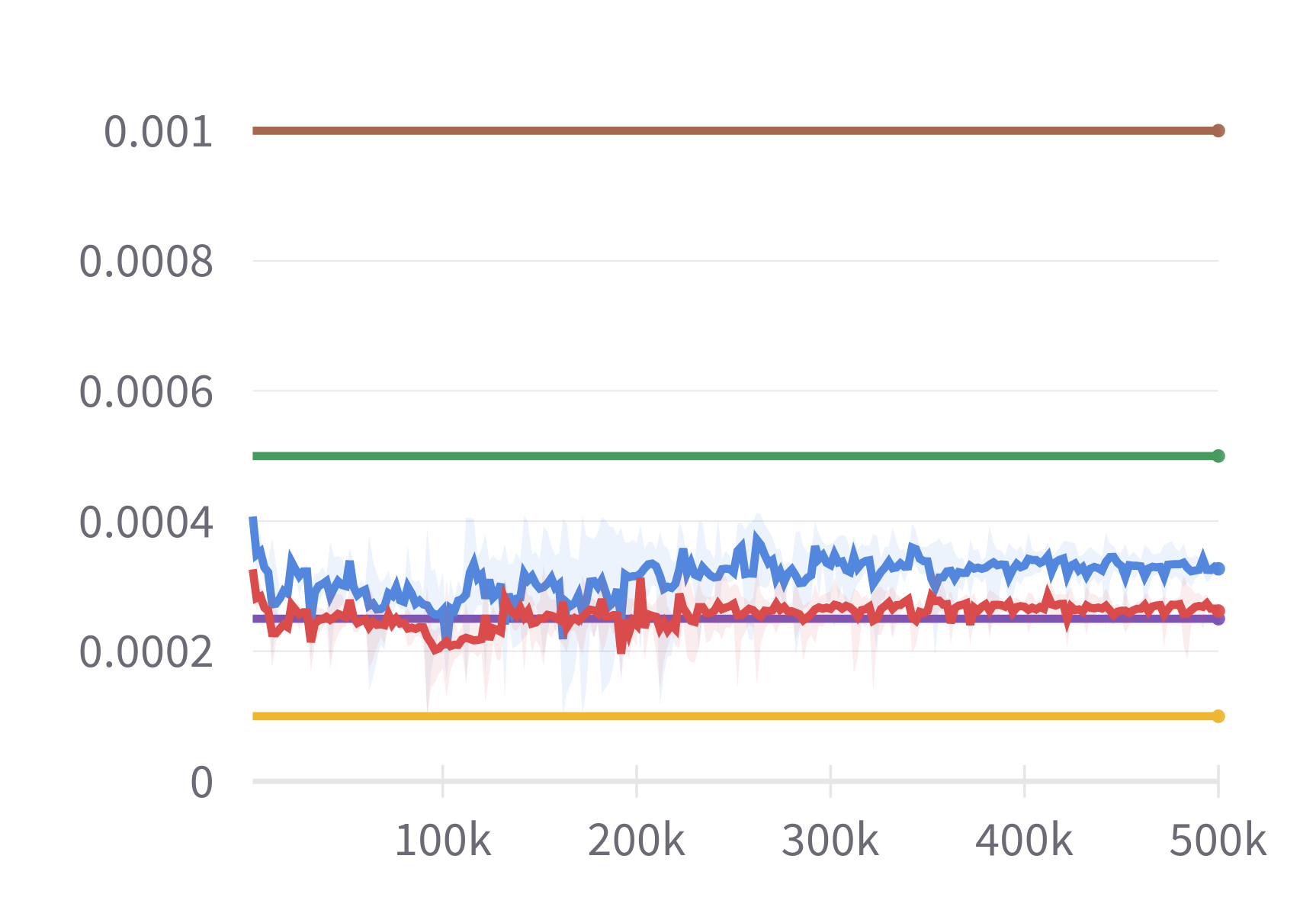}
\caption*{\small{LR-\emph{SafetyBallRun-v0}}}
\end{minipage}
}
\subfigure{
\begin{minipage}[t]{0.23\linewidth}
\centering
\includegraphics[width=1.1\textwidth]{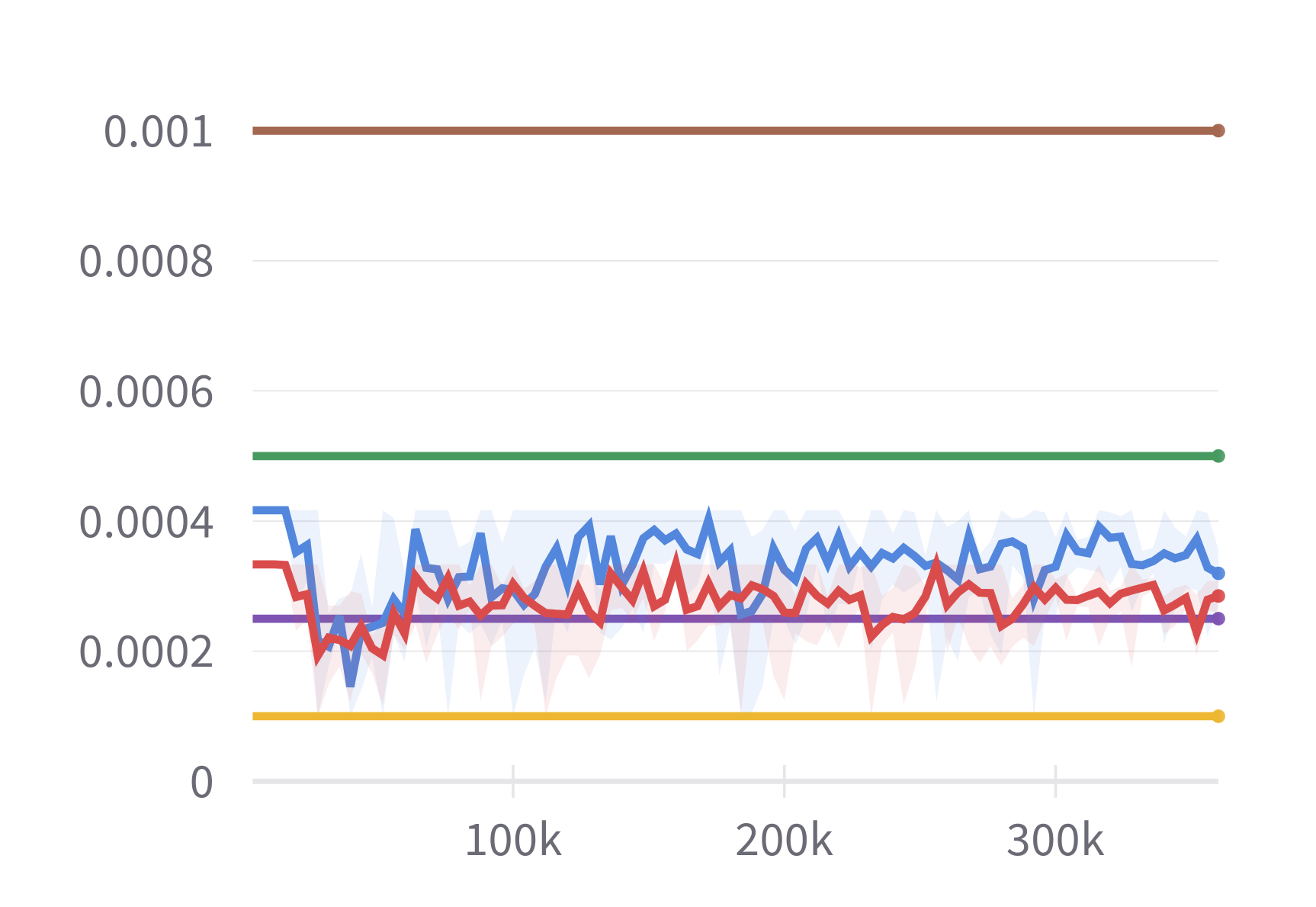}
\caption*{\small{LR-\emph{SafetyCarRun-v0}}}
\end{minipage}
}
\subfigure{
\begin{minipage}[t]{0.23\linewidth}
\centering
\includegraphics[width=1.1\textwidth]{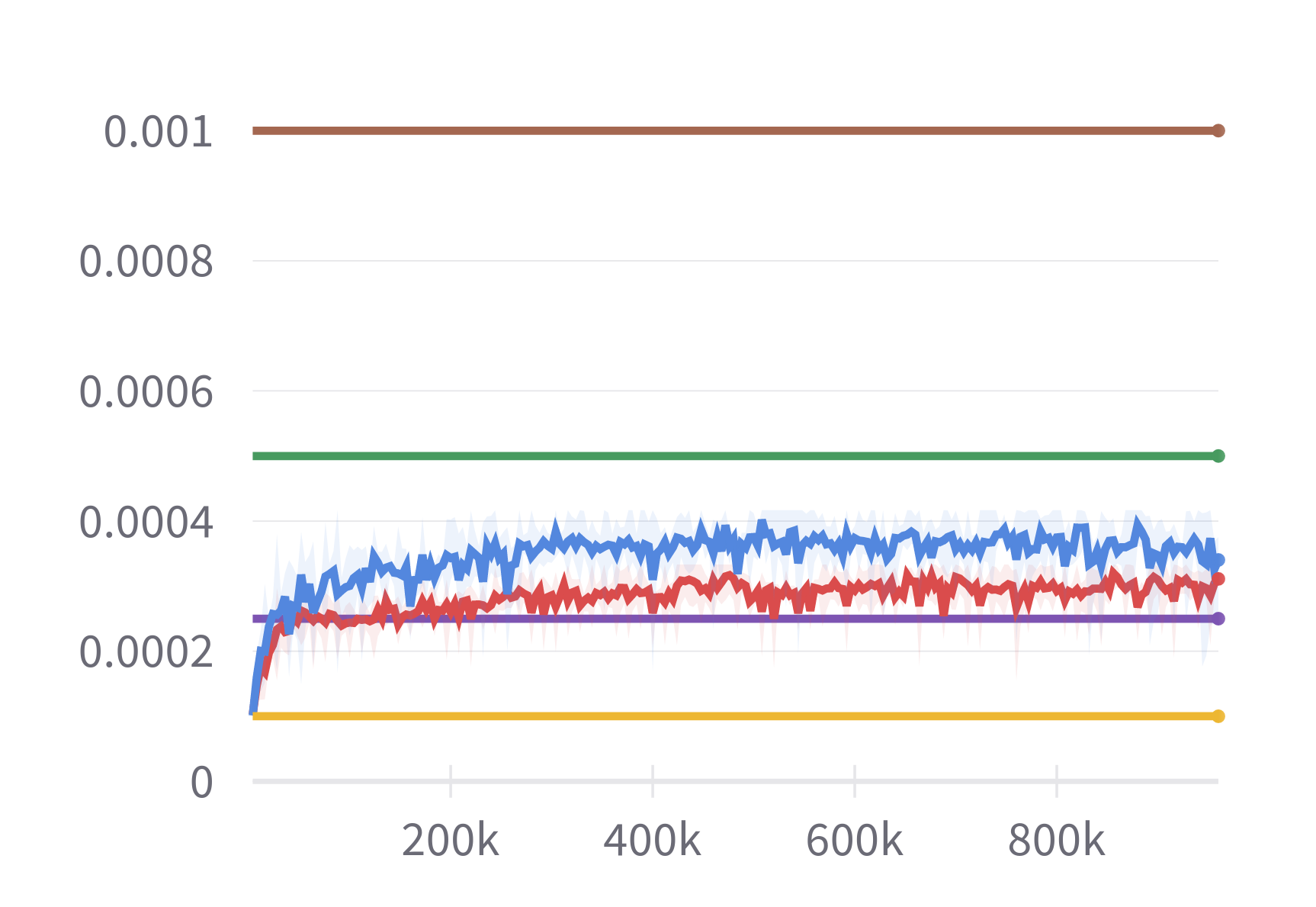}
\caption*{\small{LR-\emph{SafetyBallCircle-v0}}}
\end{minipage}
}
\subfigure{
\begin{minipage}[t]{0.23\linewidth}
\centering
\includegraphics[width=1.1\textwidth]{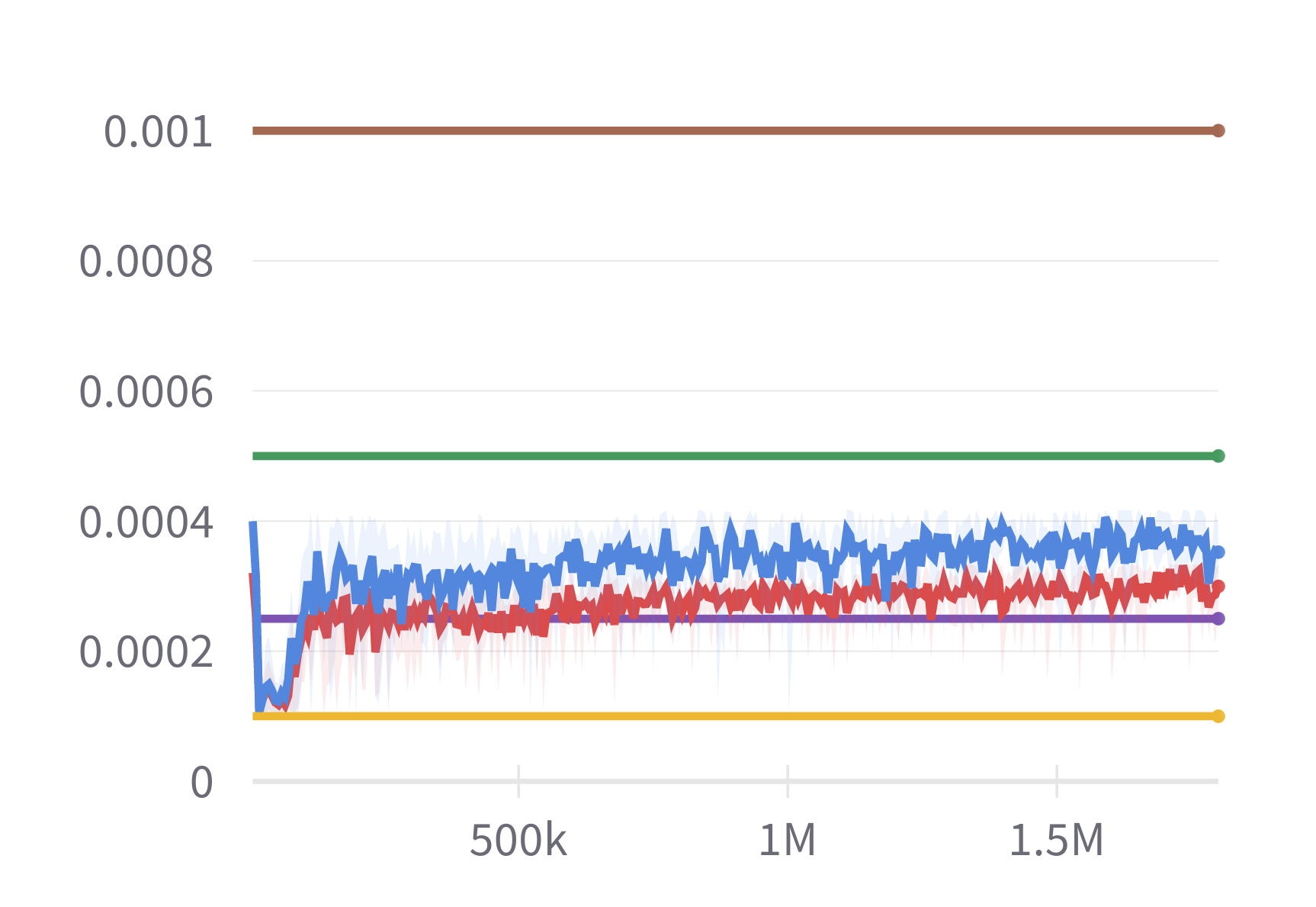}
\caption*{\small{LR-\emph{SafetyCarCircle-v0}}}
\end{minipage}
}
\centering
\caption{Learning curves for PPOL over four environments with five independent runs. In all figures, the horizontal axis is the number of time step. The solid line illustrates the mean and the shaded area depicts the maximum and the minimum. In all experiments, $H_1 = 0.001, H_2 = 3$ for \emph{InvLin}, $H_1' = 0.015, H_2' = 6$ for \emph{InvQua}, and cost limit $\textbf{d}=10$ (black dashed line).}
\label{fig_all_curves_ppol}
\end{figure*}

\section{Experiments}\label{sec_experiments}
The experimental details are deferred to the supplementary material.
\subsection{Environment}
To validate our findings, we consider the Bullet-Safety-Gym environments~\cite{gronauer2022bullet} and the Fast Safe Reinforcement Learning (FSRL) framework~\cite{liu2023datasets} in this work.
The  Bullet-Safety-Gym is a platform designed to train and evaluate safety features in constrained RL  scenarios. Meanwhile, the FSRL library offers structured modules for implementing SRL  algorithms including PPOL and DDPGL.

\renewcommand{\arraystretch}{1.5}
\begin{table}[H]
	\centering
	\fontsize{8}{8}\selectfont
	\begin{threeparttable}
		\caption{Running time (seconds) using PPOL for all experiments. Each case contains five independent runs.}
		\label{tab_running_time}
		\begin{tabular}{ccccc}
        \hline
        \hline

   			Environment & \emph{BallRun} & \emph{CarRun}
			& \emph{BallCircle} & \emph{CarCircle} \cr
        \hline
            \emph{InvLin} & $ 128.1 \pm 5.2 $ &  $ \bf{159.0 \pm 5.6} $  & $ 223.8 \pm 5.6 $ & $ \bf{861.0 \pm 28.2} $ \cr

            \emph{InvQua} & $ 128.1 \pm 2.7 $ &  $ 161.9 \pm 14.5 $  & $ 228.2 \pm 11.1 $ & $ 993.2 \pm 33.8 $ \cr

            LR=0.0001 & $ 127.1 \pm 1.5 $ &  $ 162.0 \pm 13.8 $  & $ 234.4 \pm 2.6 $ & $ 984.5 \pm 56.1 $ \cr

            LR=0.00025 & $125.8 \pm 4.3 $ &  $ 164.0 \pm 18.4 $  & $ 230.3 \pm 9.1 $ & $ 961.9 \pm 61.1 $ \cr

            LR=0.0005 & $ 124.4 \pm 3.9 $ &  $ 170.9 \pm 17.7 $  & $ 222.8 \pm 13.6 $ & $ 908.8 \pm 35.1 $ \cr

            LR=0.001 & $ \bf{120.5 \pm 6.3} $ &  $ 163.6 \pm 22.7 $  & $ \bf{214.7 \pm 14.5} $ & $ 868.9 \pm 27.9 $ \cr
   
        \hline
        \hline
		\end{tabular}
	\end{threeparttable}
\end{table}
\renewcommand{\arraystretch}{1.5}
\begin{table*}
	\centering
	\fontsize{8}{8}\selectfont
	\begin{threeparttable}
		\caption{Robustness verification for $H_1 / H_1'$ using \emph{SafetyCarRun-v0} and PPOL. Each case contains five independent runs. In all experiments, $H_2=3 / H_2'=6$ are fixed as in Figure~\ref{fig_all_curves_ppol}. The first block showcases three baselines: constant LR = 0.00025 (Baseline1), \emph{InvLin} $H_1 = 0.001$ (Baseline2), \emph{InvQua} $H_1' = 0.015$ (Baseline3), as selected in Figure~\ref{fig_all_curves_ppol}. In the rest of blocks, constant LR, $H_1$, $H_1'$ are decreased/increased by the same proportion.}
		\label{tab_robustness}
		\begin{tabular}{ccc|ccc}
        \hline
        \hline
			Parameter & Return & Cost & Parameter & Return & Cost \cr
        \hline
             Baseline1  & $538.10 \pm 8.71$ & $12.48 \pm 10.96$ &
                    
			Baseline1 $\times$ 0.2 &  $421.26 \pm 40.82$  & $9.64 \pm 4.90$ \cr
   
             Baseline2 & ${\bf 537.59 \pm 11.40}$ & ${\bf9.98 \pm 5.98}$ &
            
            Baseline2 $\times$ 0.2 & ${\bf 502.48 \pm 27.43 }$ & ${\bf 8.80 \pm 5.15}$ \cr
            
             Baseline3 & $545.22 \pm 4.93$ & $13.40 \pm 4.78$ &
            
			Baseline3 $\times$ 0.2 & $513.93 \pm 7.35$ & $11.50 \pm 4.55$ \cr

        \hline
            Baseline1 $\times$ 0.32 & $492.65 \pm 15.78$  & $7.66 \pm 5.34$ &
                    
			Baseline1 $\times$ 0.4 &  $510.41 \pm 13.62$  & $9.22 \pm 6.14$ \cr
   
            Baseline2 $\times$ 0.32 & $511.41 \pm 22.70$ & $11.46 \pm 7.22$ &
            
            Baseline2 $\times$ 0.4 & $526.35 \pm 4.30$ & $18.42 \pm 8.38$ \cr
            
            Baseline3 $\times$ 0.32 & ${\bf 509.91 \pm 29.01}$ & ${\bf 8.94 \pm 5.74}$ &
            
			Baseline3 $\times$ 0.4 & ${\bf 519.52 \pm 11.61}$ & ${\bf 8.04 \pm 5.66}$ \cr

        \hline
   			Baseline1 $\times$ 1.6 &  $545.26 \pm 4.50$  & $15.22 \pm 7.48$ &
   
            Baseline1 $\times$ 2.4 & $539.62 \pm 19.01$ & $10.02 \pm 10.87$ \cr
            
            Baseline2 $\times$ 1.6 & ${\bf 548.04 \pm 8.59}$ & ${\bf 14.32 \pm 7.21}$ &
            
            Baseline2 $\times$ 2.4 & $528.23 \pm 28.56$ & $4.74 \pm 5.68$ \cr
            
			Baseline3 $\times$ 1.6 & $ 548.45 \pm 7.52$ & $ 15.96 \pm 11.93$ & 
   
            Baseline3 $\times$ 2.4 & ${\bf 542.24 \pm 8.92}$ & ${\bf 8.28 \pm 8.27}$ \cr

        \hline
            Baseline1 $\times$ 3.2 &  $536.08 \pm 12.57$  & $6.42 \pm 8.25$ &
   
            Baseline1 $\times$ 4.0 & $529.97 \pm 15.72$ & $1.26 \pm 1.08$ \cr
            
            Baseline2 $\times$ 3.2 & $544.97 \pm 8.28$ & $10.70 \pm 5.69$ &
            
            Baseline2 $\times$ 4.0 & $ 533.70 \pm 19.98$ & $ 9.56 \pm 10.51$ \cr
            
			Baseline3 $\times$ 3.2 & ${\bf 538.08 \pm 11.93}$ & ${\bf 5.66 \pm 6.96}$ & 
   
            Baseline3 $\times$ 4.0 & ${\bf 534.07 \pm 20.06}$ & ${\bf 7.70 \pm 11.52}$ \cr
        \hline
        \hline
		\end{tabular}
	\end{threeparttable}
\end{table*}
\renewcommand{\arraystretch}{1.5}
\begin{table}[tp]
	\centering
	\fontsize{8}{8}\selectfont
	\begin{threeparttable}
		\caption{Summary of Table~\ref{tab_robustness}}
		\label{tab_robustness2}
		\begin{tabular}{cccc}
        \hline
        \hline
			Paramter & Best Performance
			& Overall Return & Overall Cost \cr
        \hline
 
			Constant LR & $536.08 / 6.42$ &  $514.17 \pm 41.45$  & $8.99 \pm 4.15$ \cr
   
            $\emph{InvLin}$-$H_1$ & $537.59 / 9.98$ & $ 529.10 \pm 15.74 $ & $10.99 \pm 4.03$ \cr
            
			$\emph{InvQua}$-$H_1'$ & ${\bf542.24 / 8.28}$ & ${\bf 531.43 \pm 14.92}$ & ${\bf 9.94 \pm 3.41}$ \cr

        \hline
        \hline
		\end{tabular}
	\end{threeparttable}
\end{table}

\subsection{Results}
We compare Algorithm~\ref{alg_papd} with the constant LR primal-dual algorithm in four environments of Bullet-Safety-Gym: \emph{SafetyBallRun-v0}, \emph{SafetyCarRun-v0}, \emph{SafetyBallCircle-v0}, \emph{SafetyCarCircle-v0} (described in the supplementary material). For a fair comparison, we maintain uniformity in all parameters and hyper-parameters (except for the LR) across each case (see the supplementary material for details of the hyper-parameters). While Algorithm~\ref{alg_papd} relies on five hyper-parameters (other than $\theta_0$), its performance is fairly robust to the choice of these parameter values. We employ the default values of $K_P, K_I, K_D$ in FSRL across all experiments. In addition, we numerically substantiate the robustness of the algorithm performance against variations in $H_1 ( H_1'), H_2 (H_2')$ values in Section~\ref{subsec_robust}.

Figure~\ref{fig_all_curves_ppol} depicts the training curves of return, cost, and LR using PPOL over five random seeds. The solid line illustrates the mean and the shaded area depicts the minimum and maximum values across seeds. More specifically, the smallest constant LR (LR = 0.0001) has the worst performance in all four environments. Despite the stable training process, this LR makes significant sacrifices in both aspects of convergence rate and optimal value (return). To some extent, the above issue can be alleviated by using a larger LR. The purple curves in Figure~\ref{fig_all_curves_ppol} show a better performance achieved by LR = 0.00025. Nevertheless, the performance is not maintained as one keeps increasing LR. Indeed, the experiments in \emph{SafetyBallCircle-v0} present that the training processes become more unstable (larger variance) and/or converge to a worse solution when LR = 0.0005 is selected. Ultimately, if LR is continuously raised until it reaches 0.001, all experiments exhibit either a significant fluctuation in return and cost, or in some cases, an even worse average performance compared to using a LR of 0.0005.

On the contrary, PAPD with \emph{InvLin} and \emph{InvQua} outperform all constant-LR cases in \emph{SafetyBallCircle-v0} and \emph{SafetyCarCircle-v0}, and achieve comparable performance in terms of return and cost with the best constant-LR trials in \emph{SafetyBallRun-v0} and \emph{SafetyCarRun-v0}. This stems from the common criterion in the optimization literature, where the solutions with higher returns but violating constraints (infeasible) are regarded as having ``inferior performance''. Moreover, we present the running time employing PPOL for all experiments, as detailed in  Table~\ref{tab_running_time}, where \emph{BallRun}, \emph{CarRun}, \emph{BallCircle}, \emph{CarCircle} serve as succinct abbreviations for the corresponding environments, namely \emph{SafetyBallRun-v0}, \emph{SafetyCarRun-v0}, \emph{SafetyBallCircle-v0}, \emph{SafetyCarCircle-v0}. Table~\ref{tab_running_time} unveils the absence of substantial difference in running time between our PAPD algorithm and the constant-LR baselines. In addition, it is noteworthy that we use the same hyper-parameters $H_1= 0.001, H_2=3$ in \emph{InvLin} and same $H_1' = 0.015, H_2' = 6$ in \emph{InvQua} across all experiments.

We also apply DDPGL to update policy parameters in the PAPD algorithm. Similar to what we observe in Figure~\ref{fig_all_curves_ppol}, \emph{InvLin} and \emph{InvQua} employing DDPGL surpasses or matches the best performance of all constant-LR cases. Given the limited space, the results of DDPGL experiments are meticulously detailed and analyzed in the supplementary material. Likewise, we maintain uniformity in hyper-parameter values throughout all conducted experiments. This consistency, despite the varied experimental scenarios, speaks to the robustness of our PAPD approach. Further evidence supporting this robustness statement will be discussed in the next subsection.

\renewcommand{\arraystretch}{1.5}
\begin{table}[tp]
	\centering
	\fontsize{6.5}{7}\selectfont
	\begin{threeparttable}
		\caption{Robustness verification for $H_2 / H_2'$ using \emph{SafetyCarRun-v0} and PPOL. Each case contains five independent runs. In all experiments, $H_1=0.001 /H_1'=0.015$ are fixed as in Figure~\ref{fig_all_curves_ppol}. The two baselines, denoted as \text{BS1} and \text{BS2}, correspond to \emph{InvLin} $H_2=3$ and \emph{InvQua} $H_2'=6$ as selected in Figure~\ref{fig_all_curves_ppol}. Starting from the first block, $H_2 /H_2'$ are increased by $10\%$.}
		\label{tab_robustness_H2}
		\begin{tabular}{ccc|ccc}
        \hline
        \hline
			$H_2 / H_2'$ & Return & Cost & $H_2 / H_2'$ & Return & Cost \cr
 
        \hline
             BS1 $\times$ 0.8 & $543.05 \pm 10.21$ & $12.40 \pm 8.66$ &

            BS1 $\times$ 0.9 & $531.73 \pm 12.61$ & $5.80 \pm 6.15$ \cr
            
            BS2 $\times$ 0.8 & $ {\bf 545.09 \pm 3.82} $ & ${\bf 6.56 \pm 5.05} $ &
              
			BS2 $\times$ 0.9 & $543.58 \pm 6.84$ & $8.04 \pm 5.26$ \cr
        \hline
            BS1 $\times$ 1.1 & $ {\bf 538.27 \pm 5.70} $ & $ {\bf 8.86 \pm 6.90} $ &

            BS1 $\times$ 1.2 & $531.27 \pm 8.97$ & $7.90 \pm 6.14$ \cr
            
            BS2 $\times$ 1.1 & $547.12 \pm 2.44$ & $13.28 \pm 2.49$ &
   
            BS2 $\times$ 1.2 & $537.37 \pm 7.68$ & $9.50 \pm 4.40$ \cr
        \hline
        \hline
		\end{tabular}
	\end{threeparttable}
\end{table}

\subsection{Robustness Verification}\label{subsec_robust}

Having observed the sensitivity of experimental results with respect to the constant-LR, one might naturally find themselves intrigued by the sensitivity exhibited by $H_1 ( H_1'), H_2 (H_2')$. 
As shown in \eqref{eqn_two_lr_prac}, $H_1 (H_1')$ plays a similar role to constant-LR in a natural way. Therefore, a compelling point of interest would involve their comparison with the constant-LR. On the other hand, $H_2 (H_2')$ is added to the LM $\lambda_k$. Indeed, for large values of $\lambda_k$, $H_2 (H_2')$ becomes more negligible, whereas for small values of $\lambda_k$, the LR  will focus on $H_2 (H_2')$ itself. The LM describes the level of complexity involved in solving the problem, which indicates that $H_2 (H_2')$ generally depends on the problems/tasks. In this subsection, we substantiate the robustness of $H_1 (H_1')$ and $H_2 ( H_2')$, respectively. The results are summarized in Tables \ref{tab_robustness}, \ref{tab_robustness2} and \ref{tab_robustness_H2}.

Table~\ref{tab_robustness} summarizes the experimental results of PPOL algorithm in \emph{SafetyCarRun-v0}, where each case contains five independent runs. More concretely, the parameters in the first block are the same as in Figure~\ref{fig_all_curves_ppol}, where constant LR (0.00025) has the most comparable performance with \emph{InvLin} ($H_1 = 0.001$) and \emph{InvQua} ($H_1' = 0.015$), and we thus select it for robustness verification. In the next three blocks, the constant LR, $H_1$, and $H_1'$ are reduced by the same proportion, and we also increase the three parameters by the same scale in the last four blocks. These 8 blocks contain a wide range of constant LR, $H_1$, $H_1'$, and thus enable us to fairly compare their sensitivity. 

The goal of problem \eqref{eqn_SRL_problem} is to maximize the expected return while satisfying the constraint, i.e., $\text{cost} < 10$ in our experiments. With this in mind, Table~\ref{tab_robustness} shows that \emph{InvLin} achieves the best performance in the first, second, and fifth blocks, while \emph{InvQua}
stands out as the epitome of exceptional performance in the rest of blocks. The fifth block is the specialty, where all three cases are infeasible. In this case, \emph{InvLin} is selected to be the best due to the smallest constraint violations and almost the largest return. To summarize, our proposed \emph{InvLin} and \emph{InvQua} outperform the constant LR in a wide range of values.

We further analyze Table~\ref{tab_robustness} from the perspective of all blocks. Indeed, Table~\ref{tab_robustness2} extracts the best performance of three parameters across all blocks.
All of them attain feasible solutions, with \emph{InvQua} emerging as the optimal choice by achieving $\text{return}=542.24, \text{cost}=8.28$. On the other hand, we collect the mean values of the return and the cost in each block of Table~\ref{tab_robustness}, and compute their mean and standard deviation across all eight blocks. These are shown in the last two columns: ``Overall Return'' and ``Overall Cost''. Notice that the best ``overall performance'', i.e., the best overall return-cost pair goes to \emph{InvQua} again. In addition, \emph{InvLin} and \emph{InvQua} yield smaller standard deviation than the constant LR case in terms of both return and cost.

Furthermore, Table~\ref{tab_robustness_H2} tests the robustness of $H_2 (H_2')$. We also use the same $H_1 (H_1'), H_2 (H_2')$ as in Figure~\ref{fig_all_curves_ppol} as our baselines. Then, we increase $H_2 (H_2')$ by $10\%$ starting from the first block in Table~\ref{tab_robustness_H2}, while keeping $H_1 (H_1')$ constant. Note that we maintain the satisfactory performance in three of the blocks and encounter small loss in the worst case while changing $H_2 (H_2')$. Moreover, \emph{InvQua} shows larger sensitivity to $H_2'$, which could be naturally explained by its quadratic format.

In summary, we are able to employ the same $H_1 (H_1'), H_2 (H_2')$ in four different  Bullet-Safety-Gym environments for both PPOL and DDPGL algorithms. In addition, Tables~\ref{tab_robustness} and \ref{tab_robustness2} indicate that \emph{InvLin} and \emph{InvQua} outperform the constant LR baselines in a wide range of $H_1 (H_1')$ that are selected, and Table~\ref{tab_robustness_H2} validates the robustness of $H_2 (H_2')$ as well. Therefore, these numerical results substantiate the robustness of our proposed PAPD algorithm.

\section{Concluding Remarks}
In this work, we propose the Adaptive Primal-Dual (APD) algorithm and its practical version (PAPD) that leverage adaptive learning rates for safe reinforcement learning (SRL). Theoretically, we provide the analyses of the APD algorithm in terms of the convergence, optimality and feasibility. We also numerically evaluate the PAPD algorithm across four well-known SRL environments in the Bullet-Safety-Gym. Our experiments show that the PAPD algorithm outperforms the primal-dual algorithm with constant learning rates. In addition, we validate the robustness of our proposed algorithm by testing across various environments, different RL methods, and a wide range of hyper-parameters associated with the two adaptive learning rates.

\section*{Acknowledgments}
This work is supported by the Rensselaer-IBM AI Research Collaboration. The authors thank Zuxin Liu for his thoughtful suggestions and valuable insights.



\clearpage

\bibliographystyle{unsrt}
\bibliography{sample}


\clearpage

\section{Supplementary Material}
To make this work self-contained and easy to read, the supplementary material is organized as follows. In the first subsection, we showcase a table of notations for the reader's effortless tracking. In Sections \ref{subsec_supple_proof1} - \ref{subsec_supple_proof2}, we provide the detailed proofs of theorems claimed in the main body of paper. In Section \ref{subsec_supple_exp}, we present the details of experiments including environment, methodology, results, and implementation.

\subsection{Table of Notations}
We present a table of notations, as outlined in Table~\ref{tab_notation}, which serves to facilitate tracking and enhance comprehension for the readers.
\begin{table}[H]
	\centering
	\fontsize{7.5}{8}\selectfont
	\begin{threeparttable}
		\caption{Table of Notations}
		\label{tab_notation}
		\begin{tabular}{|cc|cc|}
        \hline
		  \textbf{Notation} & \textbf{Definition} & \textbf{Notation} & \textbf{Definition} \cr
        \hline
 
			$S$ & state space & $A$ & action space  \cr

            $s$ & state & $a$ & action  \cr

            $R$ &  reward function & $C$ & cost function \cr 
            
            $\mathbb{P}$ &transition probability & $\gamma$ & discount factor \cr

            $\mathcal{U}$ & initial state distribution & $\pi$ & policy \cr 

            $\Theta$ & parameterization space & $\tau$ & trajectory \cr

            $\theta$ & primal variable & $\lambda$ & dual variable \cr

            $J_R$ & expected return & $J_C$ & expected cost \cr

            $B$ & cost bound & $\mathcal{L}$ & Lagrangian\cr

            LM & Lagrangian multiplier & LR & learning rate \cr
             
            $\textbf{d}$ & constraint threshold &  $d$ & dual function \cr

           $\eta_k$ & primal stepsize & $\zeta$ & dual stepsize \cr

           $\lambda^*$ & optimal dual variable & $D^*$ & dual optimum  \cr

           $g$ & constraint function & $\epsilon_k$ & Lagrangian error \cr

           $\mu$ & strongly convex constant & $L$ & Lipschitz constant \cr

           $H_1(H_2)$ & practical LR constant & $K_P$ & proportional gain \cr

           $K_I$ & integral gain & $K_D$ & derivative gain \cr

        \hline
		\end{tabular}
	\end{threeparttable}
\end{table}

\subsection{Proof of Theorem~\ref{theorem_dual_convergence}}\label{subsec_supple_proof1}
We start by recalling the dual update \eqref{eqn_dual_update}
\begin{align}\label{eqn_dual_update_supp}
    \lambda_{k+1} = \left[\lambda_k+ \zeta g(\theta_{k+1})\right]_+.
\end{align}
Then, we obtain
\begin{align}
     || \lambda_{k+1} - \lambda^* ||^2 &= || \left[\lambda_k+ \zeta g(\theta_{k+1}) \right]_+ - \lambda^* ||^2 \nonumber \\
    &\leq || \lambda_k- \lambda^* +\zeta g(\theta_{k+1}) ||^2,
\end{align}
where the last inequality follows from the non-expansiveness of the projection.

Expanding the norm square yields
\begin{align}
    || \lambda_{k+1} - \lambda^* ||^2  &\leq || \lambda_k- \lambda^* ||^2 + \zeta^2 || g(\theta_{k+1}) ||^2 \nonumber \\
    &\quad +2 \zeta (\lambda_{k} - \lambda^*)^T g(\theta_{k+1}). 
\end{align}
Combining Assumption~\ref{assumption_bound_cost} and \eqref{eqn_performance_measure_cost} yields $|| J_C || \leq B / (1-\gamma)$. By the definition of $g(\theta_{k+1})$ in \eqref{eqn_constraint_func} and the triangle inequality we obtain
\begin{align}\label{eqn_g_norm_square}
    ||g(\theta_{k+1})||^2 & \leq (|| J_C || + || {\bf d} ||)^2 \leq ( \frac{B}{1-\gamma} + || {\bf d}|| )^2.
\end{align}

Then we have
\begin{align}\label{eqn_convergence_before_lemma}
    || \lambda_{k+1} - \lambda^* ||^2  &\leq || \lambda_k- \lambda^* ||^2 + \zeta^2 ( \frac{B}{1-\gamma} + || {\bf d}|| )^2 \nonumber \\
    &\quad + 2 \zeta (\lambda_{k} - \lambda^*)^T g(\theta_{k+1}). 
\end{align}
To proceed, we rely on the following technical lemma.
\begin{lemma}
\label{lemma_primal_error}
Consider the dual function $d(\cdot)$ defined in \eqref{eqn_dual_function}. $\forall \lambda_1, \lambda_2 \in \Real^m$, denote by $\theta^*(\lambda_1)$ and $\theta^*(\lambda_2)$ the optimizer of the Lagrangian with respect to $\lambda_1$ and $\lambda_2$. Suppose that there exists a $\epsilon > 0$ and a $\theta^{\dagger} (\lambda_1)$ such that
\begin{align}\label{eqn_primaL_err}
     \mathcal{L}(\theta^\dagger(\lambda_1), \lambda_1) - \mathcal{L}(\theta^*(\lambda_1), \lambda_1) = \epsilon.
\end{align}
Then, it holds that
\begin{align}
    d(\lambda_2) \leq d(\lambda_1) +  (\lambda_2 - \lambda_1)^T g(\theta^\dagger(\lambda_1)) + \epsilon.
\end{align}
\end{lemma}
\begin{myproof}
We proceed by recalling the definition of the dual function $d(\lambda_1)$ and $d(\lambda_2)$ and computing their difference
\begin{align}\label{eqn_dual_difference}
    d(\lambda_1) - d(\lambda_2) = \mathcal{L}(\theta^*(\lambda_1), \lambda_1) - \mathcal{L}(\theta^*(\lambda_2), \lambda_2),
\end{align}

By virtue of the optimality of $\theta^*(\lambda_2)$, \eqref{eqn_dual_difference} can be rewritten as
\begin{align}
    d(\lambda_1) - d(\lambda_2) \geq \mathcal{L}(\theta^*(\lambda_1), \lambda_1) - \mathcal{L}(\theta^\dagger(\lambda_1), \lambda_2).
\end{align}
Substituting \eqref{eqn_primaL_err} into the previous inequality yields
\begin{align}
    d(\lambda_1) - d(\lambda_2) \geq \mathcal{L}(\theta^\dagger(\lambda_1), \lambda_1) - \mathcal{L}(\theta^\dagger(\lambda_1), \lambda_2) - \epsilon.
\end{align}
By expanding the Lagrangian the previous inequality reduces to
\begin{align}\label{eqn_dlam1_dlam2}
    d(\lambda_1) - d(\lambda_2) &\geq -J_R(\pi_{\theta^\dagger(\lambda_1)}) + \lambda_1^T g(\theta^\dagger(\lambda_1)) - \nonumber \\
    & \quad \left(-J_R(\pi_{\theta^\dagger(\lambda_1)}) + \lambda_2^T g(\theta^\dagger(\lambda_1)) \right) - \epsilon \nonumber \\
    &=(\lambda_1 - \lambda_2)^T g(\theta^\dagger(\lambda_1)) - \epsilon.
\end{align}
Reordering \eqref{eqn_dlam1_dlam2} completes the proof of Lemma~\ref{lemma_primal_error}.
\end{myproof}

Employ Lemma~\ref{lemma_primal_error} with $\lambda^*$ and $\lambda_k$ yields
\begin{align}\label{eqn_lemma_primaL_err}
    d(\lambda^*) \leq d(\lambda_k) + (\lambda^* - \lambda_k)^T g(\theta_{k+1}) + \epsilon_k,
\end{align}
where $g(\theta_{k+1}) = g(\theta^\dagger(\lambda_k))$ and $\epsilon_k$ denotes the primal error of updating the Lagrangian, as stated in Theorem~\ref{theorem_dual_convergence}.

Combing \eqref{eqn_convergence_before_lemma} and \eqref{eqn_lemma_primaL_err} yields
\begin{align}
    || \lambda_{k+1} - \lambda^* ||^2  &\leq || \lambda_k- \lambda^* ||^2 + \zeta^2 ( \frac{B}{1-\gamma} + || {\bf d}|| )^2 \nonumber \\
    &\quad + 2 \zeta (d(\lambda_k)-D^* +\epsilon_k). 
\end{align}
where $D^* = d(\lambda^*).$ Unrolling the previous inequality starting from $\lambda_K$ to $\lambda_0$ yields
\begin{align}
    0 \leq || \lambda_{K} - \lambda^* ||^2  &\leq || \lambda_0- \lambda^* ||^2 + K \zeta^2 ( \frac{B}{1-\gamma} + || {\bf d}|| )^2 \nonumber \\
    &\quad +2 \zeta (\sum_{k=0}^{K-1} d(\lambda_k)-K D^* + \sum_{k=0}^{K-1} \epsilon_k). 
\end{align}
Reordering terms in the previous expression yields
\begin{align}\label{eqn_convexity1}
    D^* \! - \! \frac{1}{K} \sum_{k=0}^{K-1}  d(\lambda_{k})
    &\leq  \frac{|| \lambda_{0}- \lambda^* ||^2 }{2 \zeta K}  + \frac{\zeta ( B + (1-\gamma)|| {\bf d}|| )^2}{2 (1-\gamma)^2} \nonumber \\
    & \quad + \frac{1}{K} \sum_{k=0}^{K-1} \epsilon_k.
\end{align}

Note that $\lambda_{\text{best}} \in \argmax_{\lambda \in \{\lambda_k\}_{k=0}^K }  d(\lambda)$, it holds that
\begin{align}
    0 \leq D^* - d(\lambda_{\text{best}})
    &\leq  \frac{|| \lambda_{0}- \lambda^* ||^2 }{2 \zeta K}  + \frac{\zeta ( B + (1-\gamma)|| {\bf d}|| )^2}{2 (1-\gamma)^2} \nonumber \\
    & \quad + \frac{1}{K} \sum_{k=0}^{K-1} \epsilon_k.
\end{align}
where the leftmost inequality follows from the definition of the dual function. This completes the proof of Theorem~\ref{theorem_dual_convergence}.


\subsection{Proof of Theorem~\ref{theorem_primal_bound}}
For $\forall k > 0$, the following chain of inequalities holds due to the fact that the dual function is a lower bound on the value of the primal and it is concave
\begin{align}\label{eqn_chain_inequality}
    -J_R(\pi_{\theta^*}) \geq d\left( \frac{1}{K} \sum_{k=0}^{K-1} \lambda_k\right) \geq \frac{1}{K} \sum_{k=0}^{K-1}  d(\lambda_k).
\end{align}

Since the primal variable $\theta_{k+1}$ is updated by the APD algorithm (Algorithm~\ref{alg_apd}) with primal error $\epsilon_k$ and by definition $d(\lambda_k) = \mathcal{L}(\theta^*(\lambda_k), \lambda_k)$, we obtain
\begin{align}
    d(\lambda_k) &= \mathcal{L}(\theta_{k+1}, \lambda_k) - \epsilon_k \nonumber \\
    &= -J_R(\pi_{\theta_{k+1}}) + \lambda_k^T g(\theta_{k+1}) - \epsilon_k.
\end{align}
Substituting the previous equation into \eqref{eqn_chain_inequality} yields
\begin{align}
    -J_R(\pi_{\theta^*})  \geq \frac{1}{K} \sum_{k=0}^{K-1}  \left( -J_R(\pi_{\theta_{k+1}}) + \lambda_k^T g(\theta_{k+1}) - \epsilon_k \right).
\end{align}
By reordering terms the previous inequality reduces to
\begin{align}\label{eqn_state_augment1}
    \frac{1}{K} \sum_{k=0}^{K-1} J_R(\pi_{\theta_{k+1}}) &\geq J_R(\pi_{\theta^*}) + \frac{1}{K} \sum_{k=0}^{K-1}  \lambda_k^T g(\theta_{k+1})  \nonumber \\
    &\quad -\frac{1}{K} \sum_{k=0}^{K-1} \epsilon_k.
\end{align}

By virtue of Lemma 1 in \cite{calvo2021state} and the fact that $||g(\theta_{k+1})||^2 \leq ( B / (1-\gamma) + || {\bf d}|| )^2$ as described in \eqref{eqn_g_norm_square}, we obtain
\begin{align}\label{eqn_state_augment2}
    \frac{1}{K} \sum_{k=0}^{K-1}  \lambda_k^T g(\theta_{k+1}) \geq -\frac{\zeta ( B + (1-\gamma)|| {\bf d}|| )^2}{2 (1-\gamma)^2} - \frac{|| \lambda^0||^2}{2 \zeta K}.
\end{align}
Combining \eqref{eqn_state_augment1} and \eqref{eqn_state_augment2} yields
\begin{align}
    \frac{1}{K} \sum_{k=0}^{K-1} J_R(\pi_{\theta_{k+1}}) &\geq J_R(\pi_{\theta^*}) -\frac{1}{K} \sum_{k=0}^{K-1} \epsilon_k  \nonumber \\
    &\quad  - \frac{\zeta ( B + (1-\gamma)|| {\bf d}|| )^2}{2 (1-\gamma)^2} - \frac{|| \lambda^0||^2}{2 \zeta K}.
\end{align}
The proof is completed by taking the limit inferior in both sides of the previous inequality. 

\begin{figure*}
\centering
\subfigure[Ball]{
\begin{minipage}[t]{0.23\linewidth}
\centering
\includegraphics[width=1\textwidth]{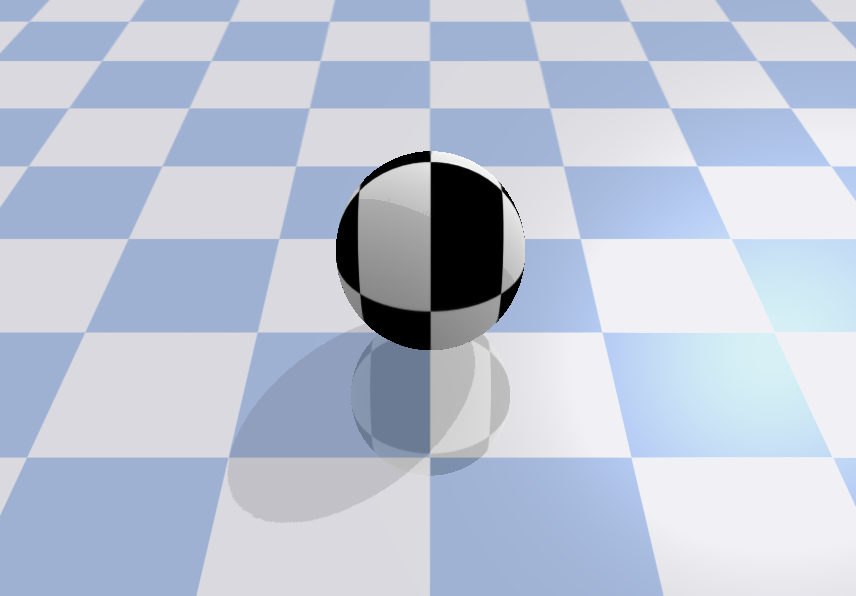}
\end{minipage}
}
\subfigure[Car]{
\begin{minipage}[t]{0.23\linewidth}
\centering
\includegraphics[width=1\textwidth]{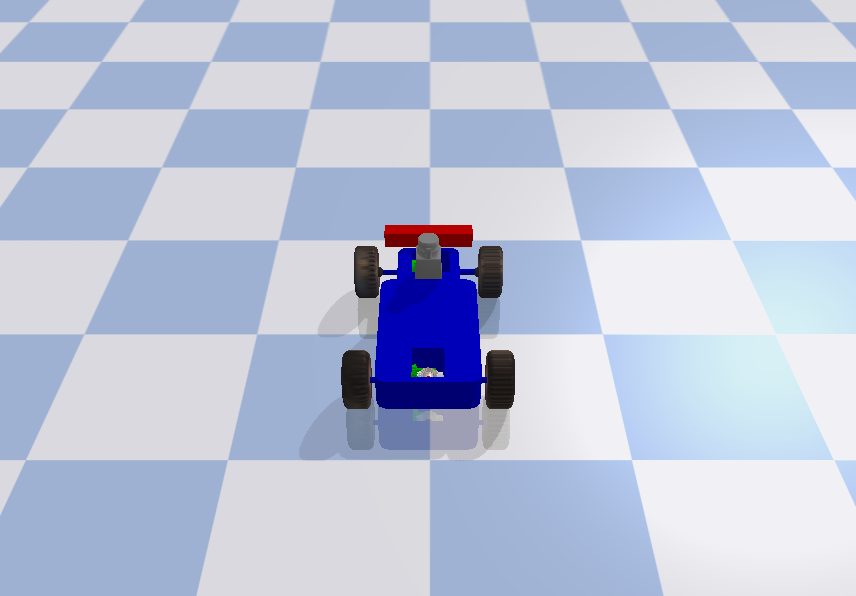}
\end{minipage}
}
\subfigure[Run]{
\begin{minipage}[t]{0.23\linewidth}
\centering
\includegraphics[width=1\textwidth]{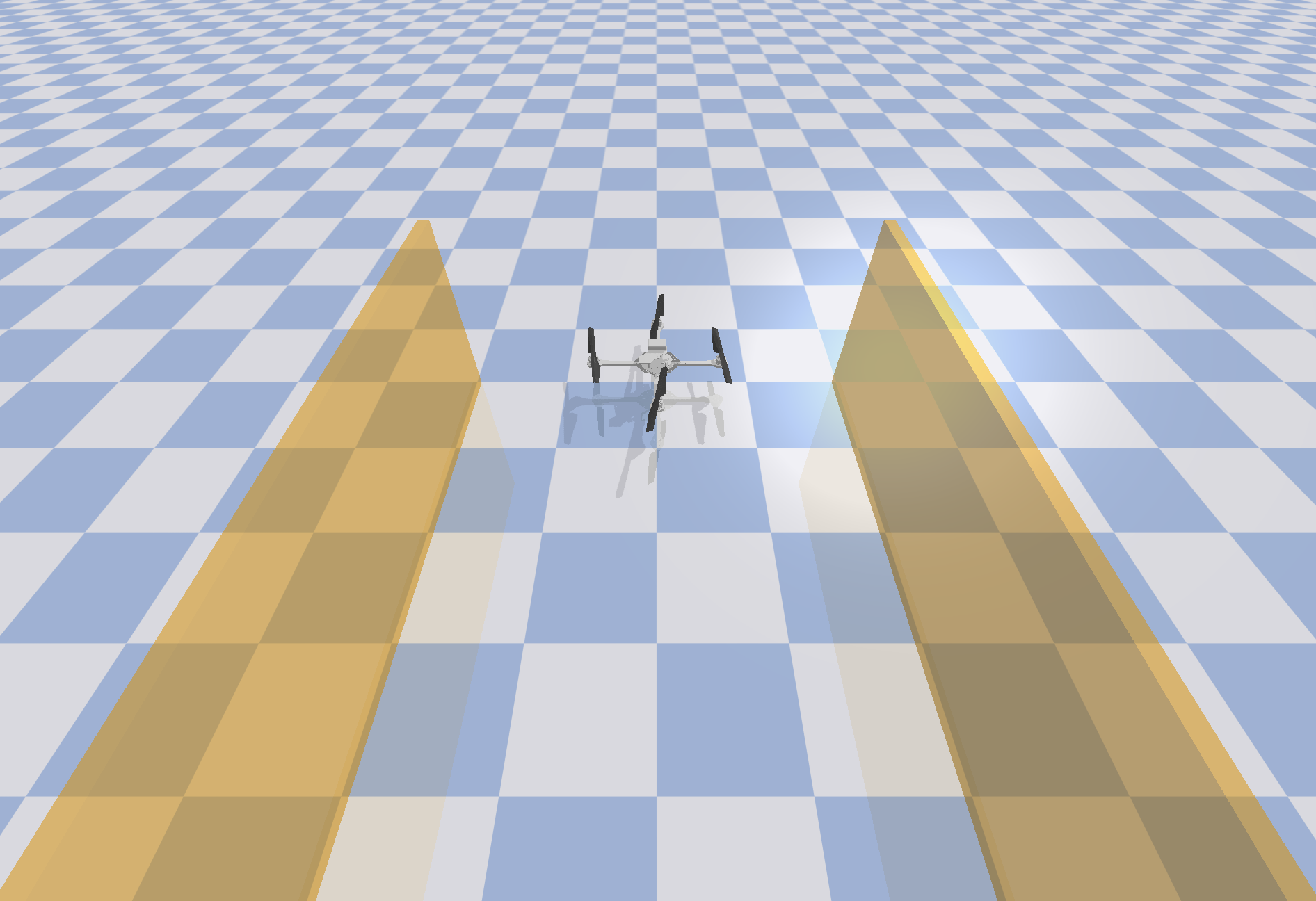}
\end{minipage}
}
\subfigure[Circle]{
\begin{minipage}[t]{0.23\linewidth}
\centering
\includegraphics[width=1\textwidth]{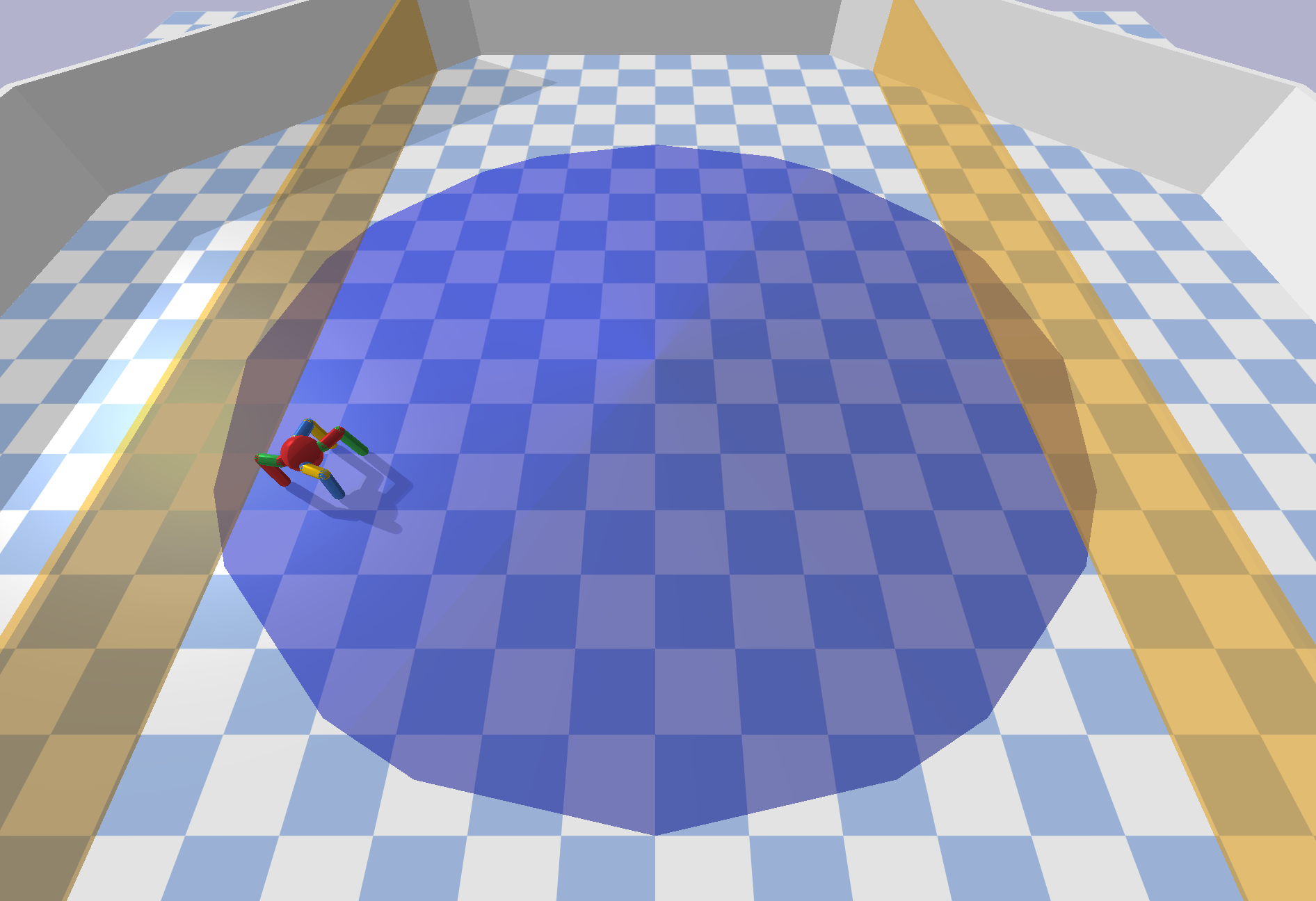}
\end{minipage}
}
\centering
\caption{Agents and Tasks from \cite{gronauer2022bullet}: (a) The Ball agent; (b) The Car agent; (c) The Run task; (d) The Circle task.}
\label{fig_illustration}
\end{figure*}
\begin{figure*}
\centering
\subfigure{
\begin{minipage}[t]{0.15\linewidth}
\centering
\includegraphics[width=0.7\textwidth]{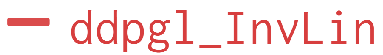}
\end{minipage}
}
\subfigure{
\begin{minipage}[t]{0.15\linewidth}
\centering
\includegraphics[width=0.7\textwidth]{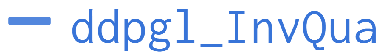}
\end{minipage}
}
\subfigure{
\begin{minipage}[t]{0.15\linewidth}
\centering
\includegraphics[width=0.7\textwidth]{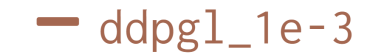}
\end{minipage}
}
\subfigure{
\begin{minipage}[t]{0.15\linewidth}
\centering
\includegraphics[width=0.7\textwidth]{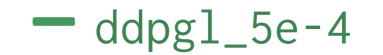}
\end{minipage}
}
\subfigure{
\begin{minipage}[t]{0.15\linewidth}
\centering
\includegraphics[width=0.7\textwidth]{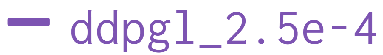}
\end{minipage}
}
\subfigure{
\begin{minipage}[t]{0.15\linewidth}
\centering
\includegraphics[width=0.7\textwidth]{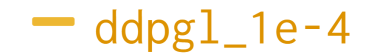}
\end{minipage}
}
\subfigure{
\begin{minipage}[t]{0.23\linewidth}
\centering
\includegraphics[width=1.1\textwidth]{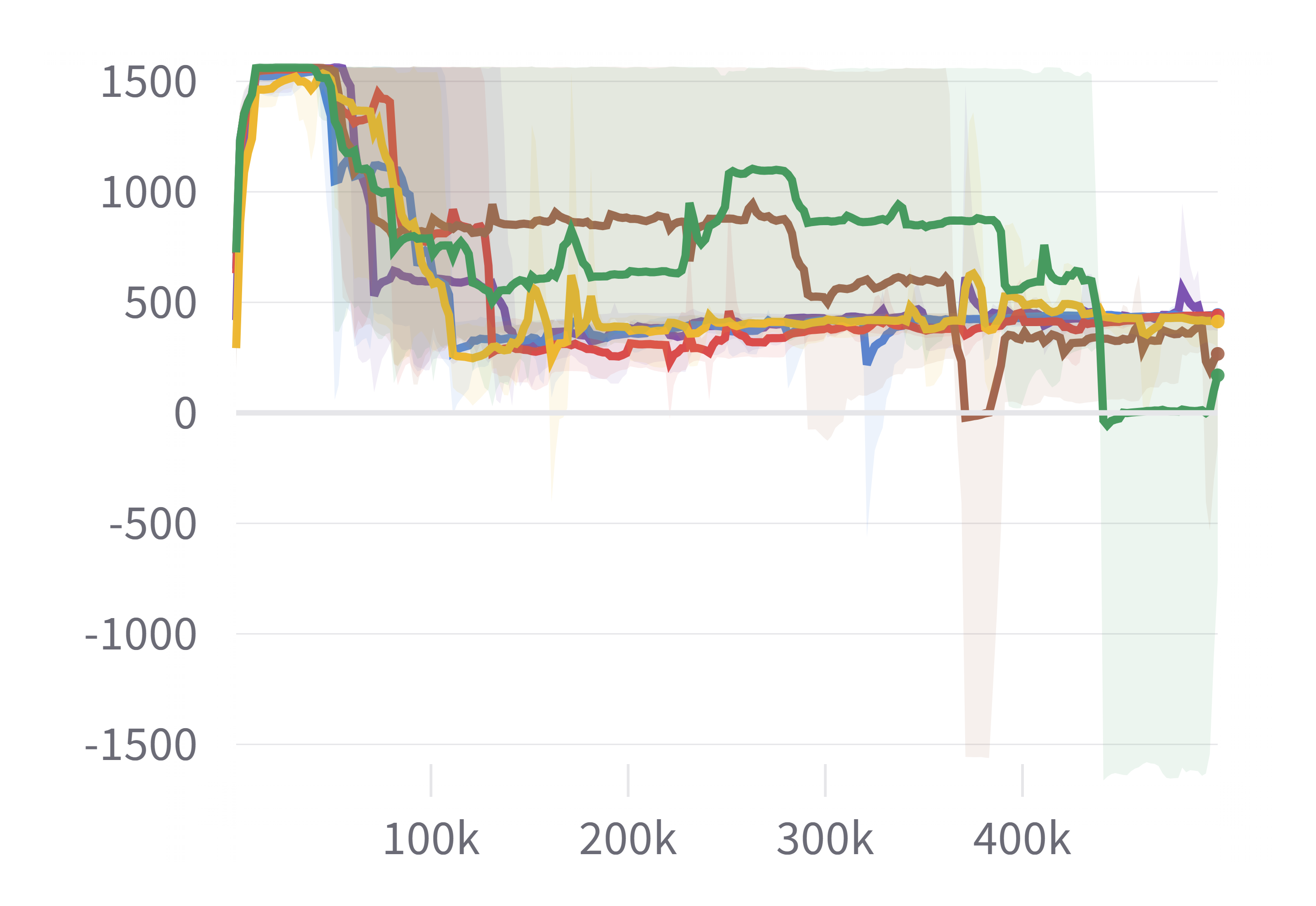}
\caption*{\small{Return-\emph{SafetyBallRun-v0}}}
\end{minipage}
}
\subfigure{
\begin{minipage}[t]{0.23\linewidth}
\centering
\includegraphics[width=1.1\textwidth]{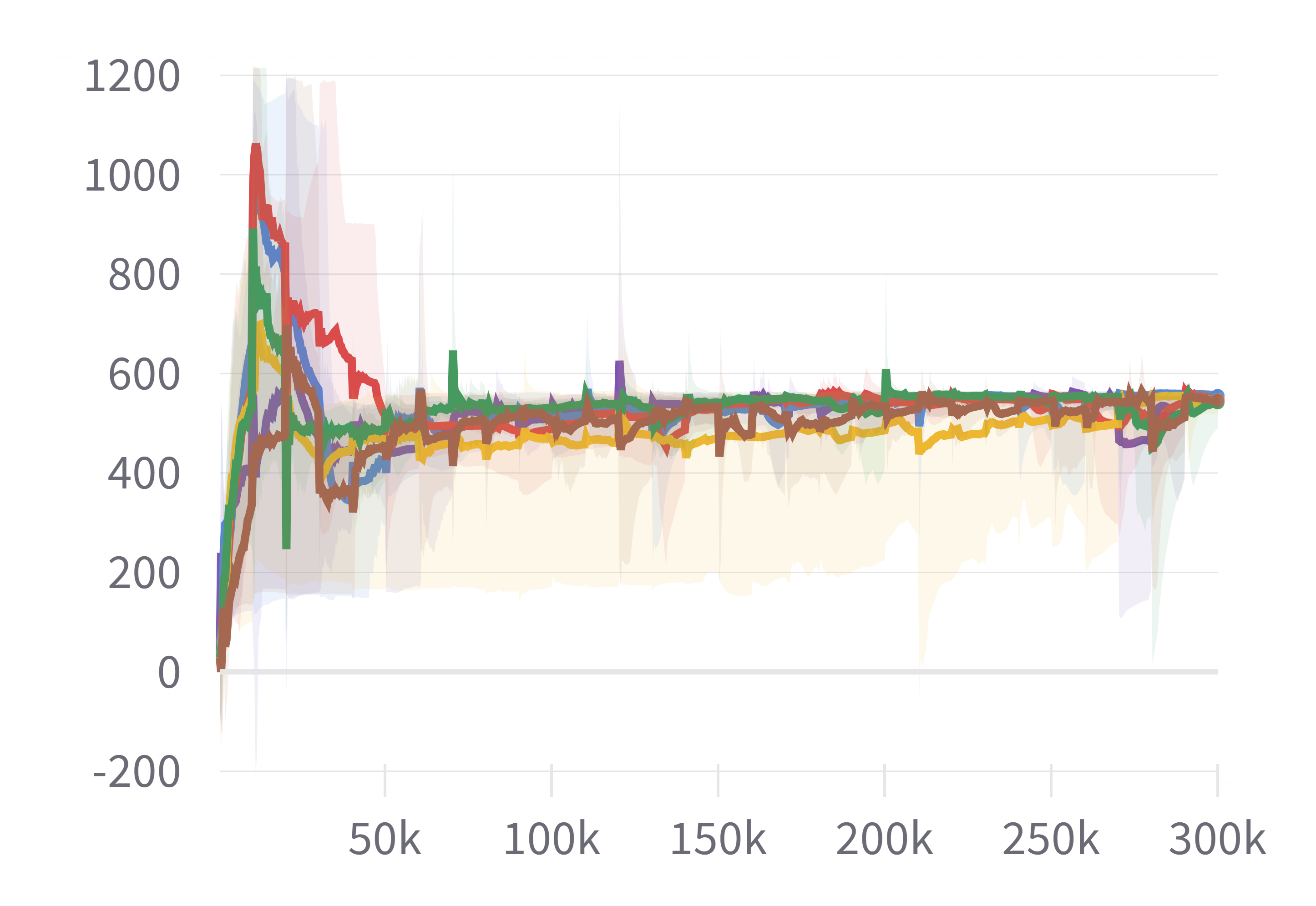}
\caption*{\small{Return-\emph{SafetyCarRun-v0}}}
\end{minipage}
}
\subfigure{
\begin{minipage}[t]{0.23\linewidth}
\centering
\includegraphics[width=1.1\textwidth]{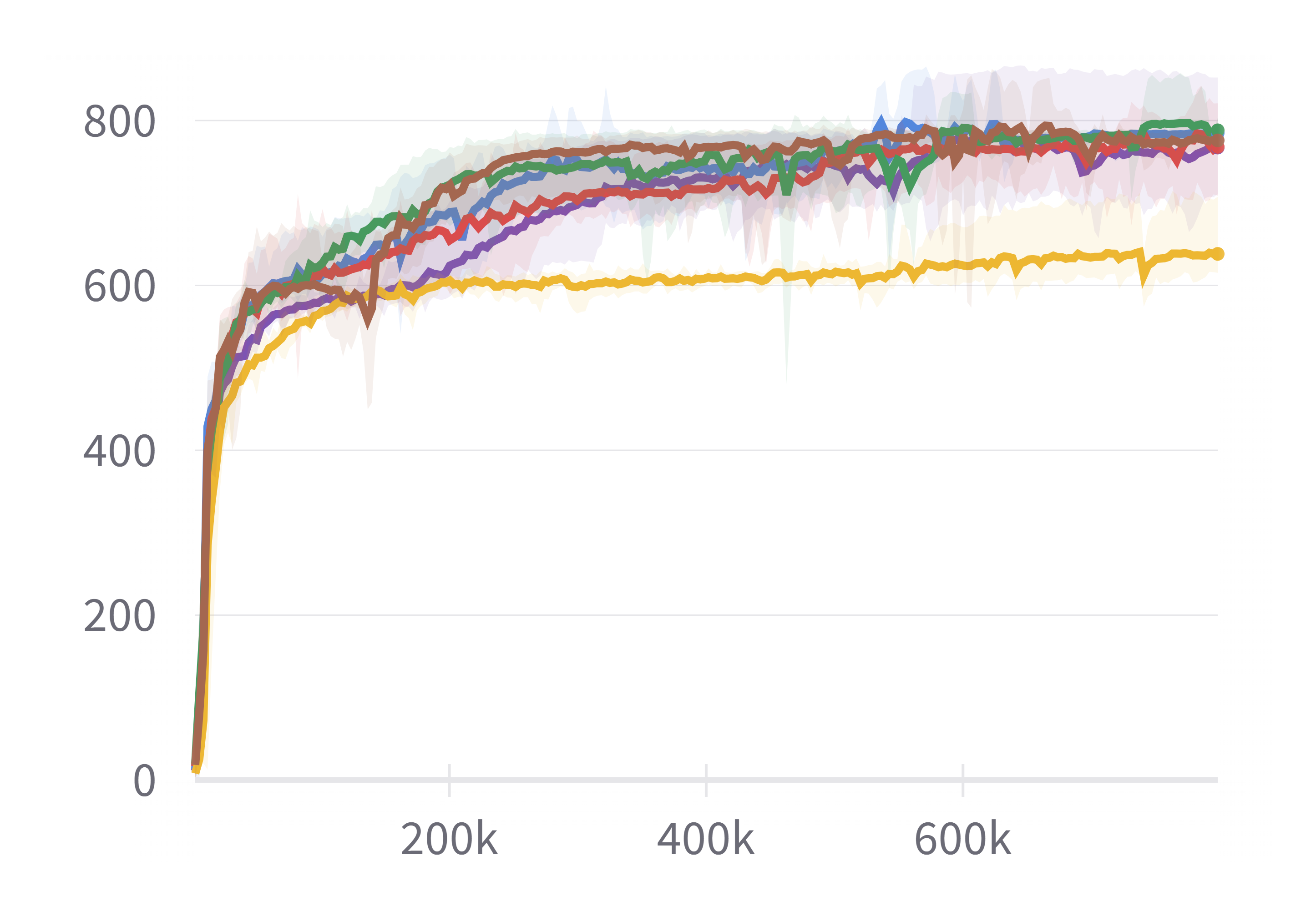}
\caption*{\small{Return-\emph{SafetyBallCircle-v0}}}
\end{minipage}
}
\subfigure{
\begin{minipage}[t]{0.23\linewidth}
\centering
\includegraphics[width=1.1\textwidth]{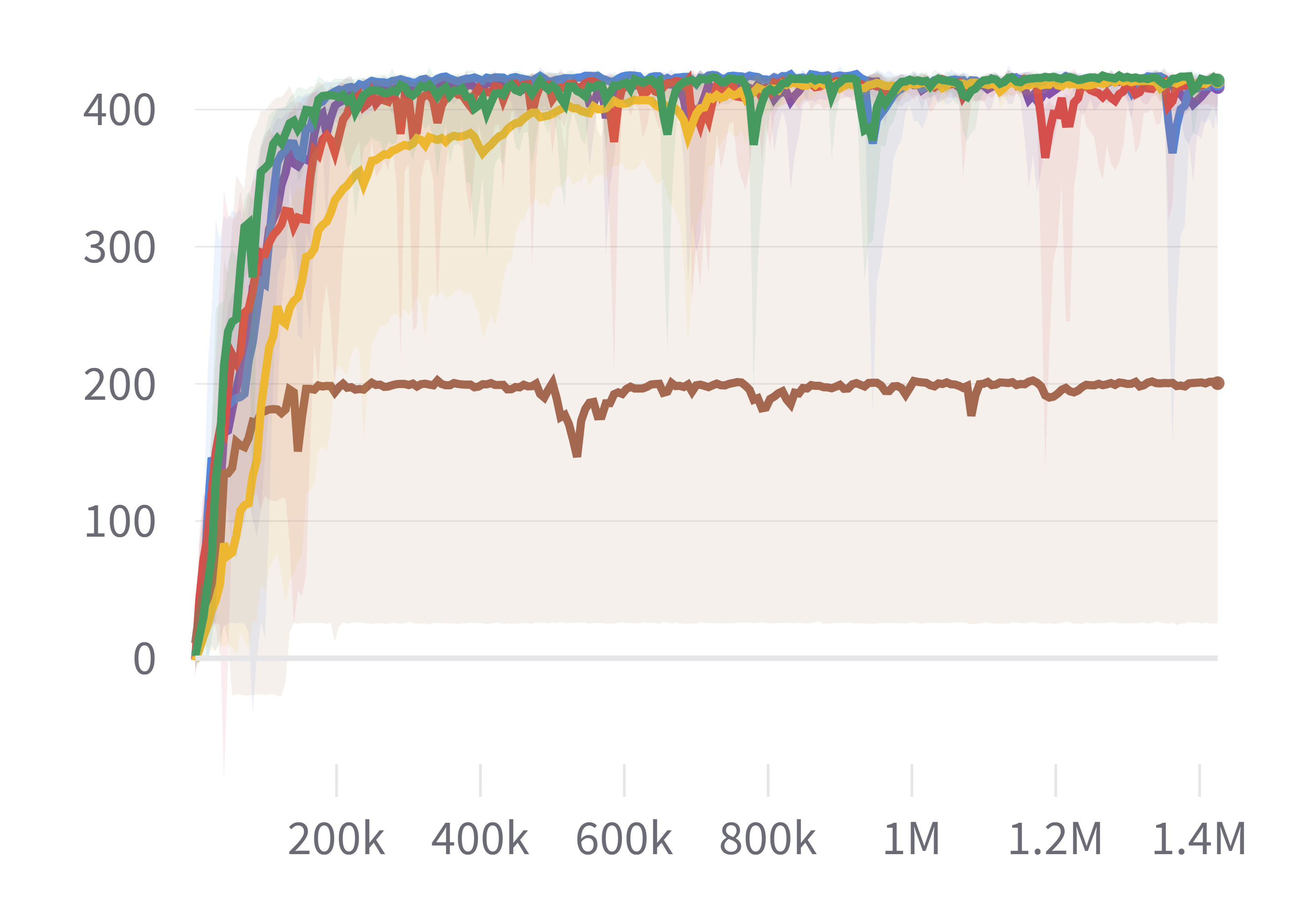}
\caption*{\small{Return-\emph{SafetyCarCircle-v0}}}
\end{minipage}
}
\subfigure{
\begin{minipage}[t]{0.23\linewidth}
\centering
\includegraphics[width=1.1\textwidth]{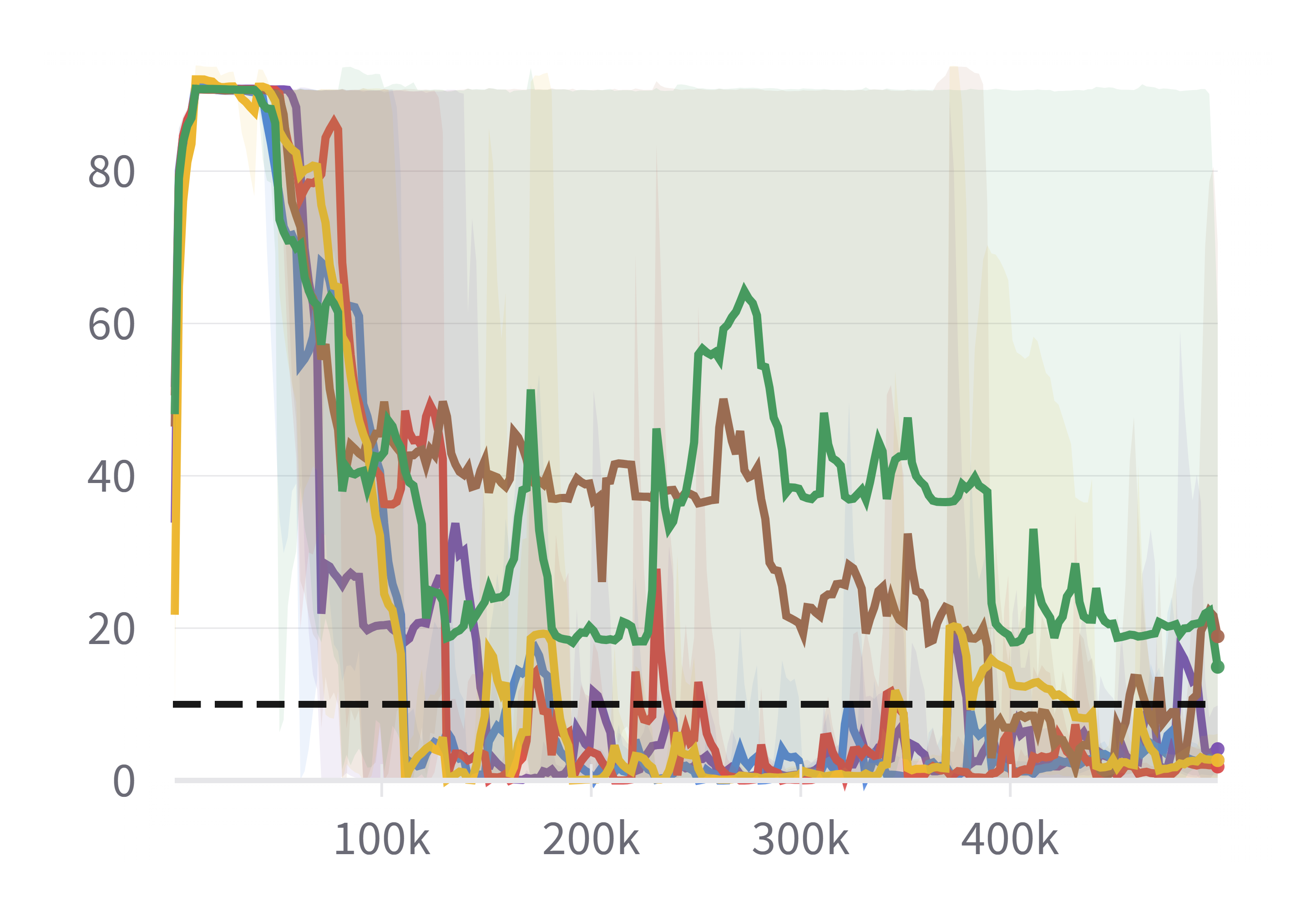}
\caption*{\small{Cost-\emph{SafetyBallRun-v0}}}
\end{minipage}
}
\subfigure{
\begin{minipage}[t]{0.23\linewidth}
\centering
\includegraphics[width=1.1\textwidth]{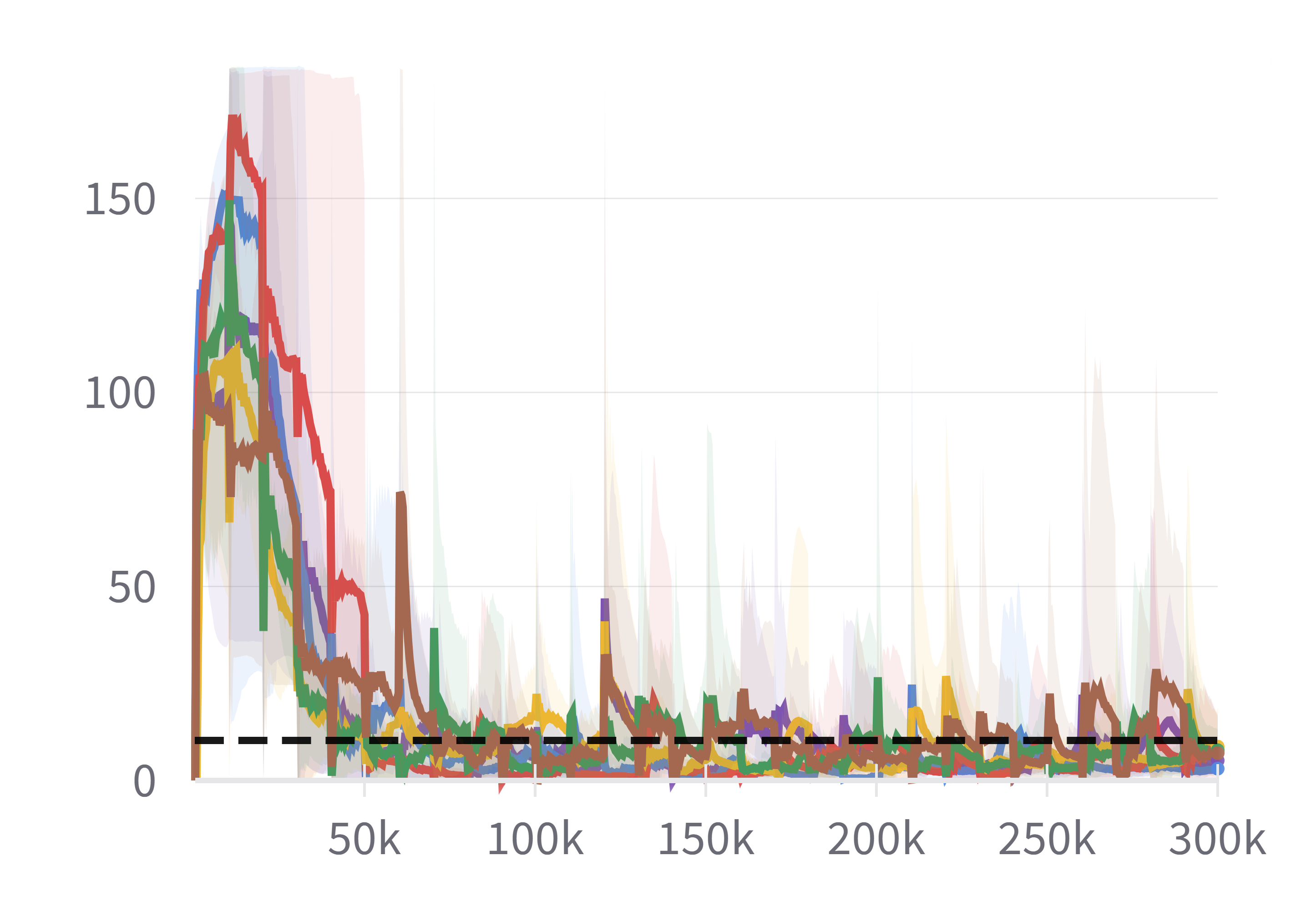}
\caption*{\small{Cost-\emph{SafetyCarRun-v0}}}
\end{minipage}
}
\subfigure{
\begin{minipage}[t]{0.23\linewidth}
\centering
\includegraphics[width=1.1\textwidth]{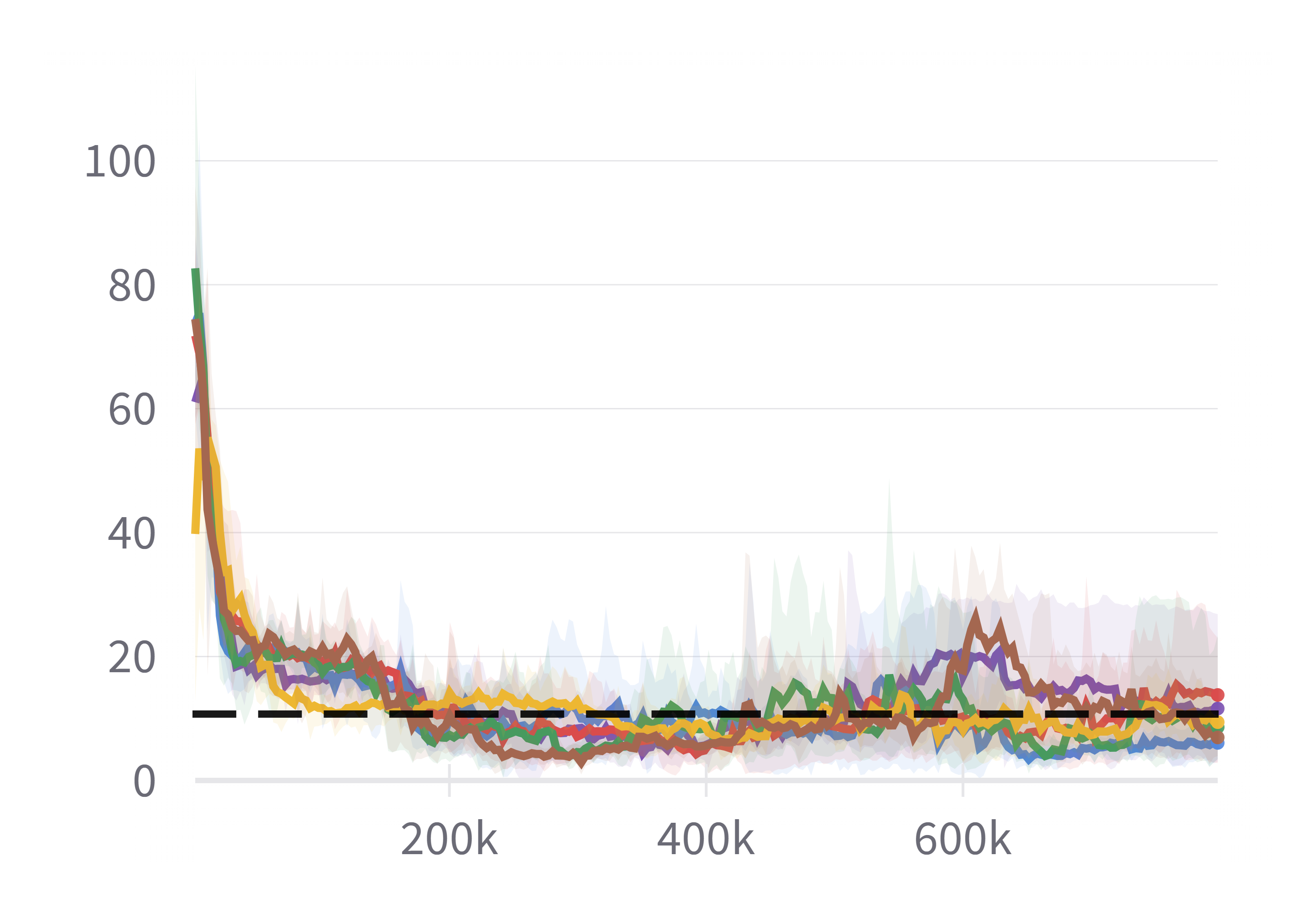}
\caption*{\small{Cost-\emph{SafetyBallCircle-v0}}}
\end{minipage}
}
\subfigure{
\begin{minipage}[t]{0.23\linewidth}
\centering
\includegraphics[width=1.1\textwidth]{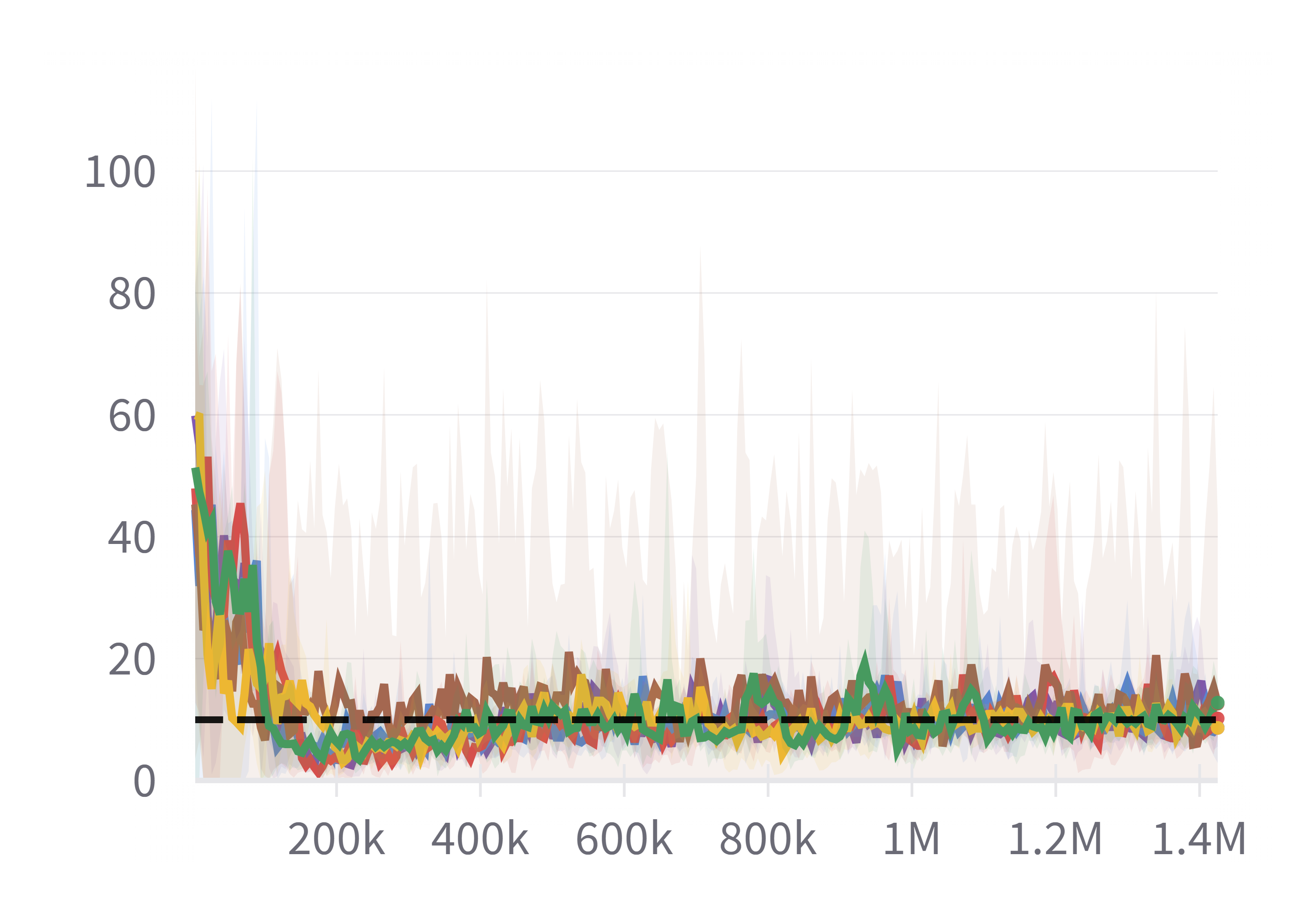}
\caption*{\small{Cost-\emph{SafetyCarCircle-v0}}}
\end{minipage}
}
\subfigure{
\begin{minipage}[t]{0.23\linewidth}
\centering
\includegraphics[width=1.1\textwidth]{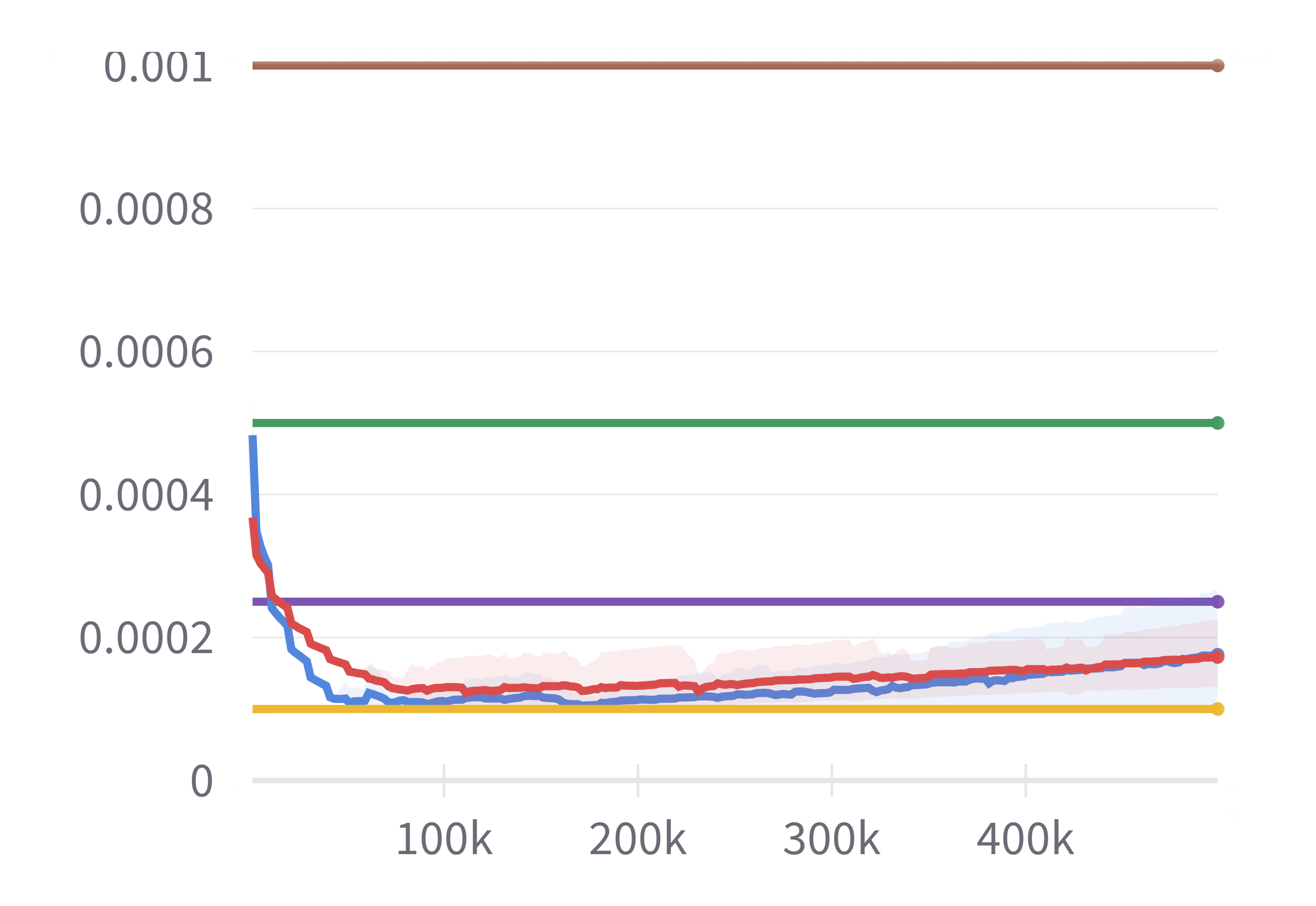}
\caption*{\small{LR-\emph{SafetyBallRun-v0}}}
\end{minipage}
}
\subfigure{
\begin{minipage}[t]{0.23\linewidth}
\centering
\includegraphics[width=1.1\textwidth]{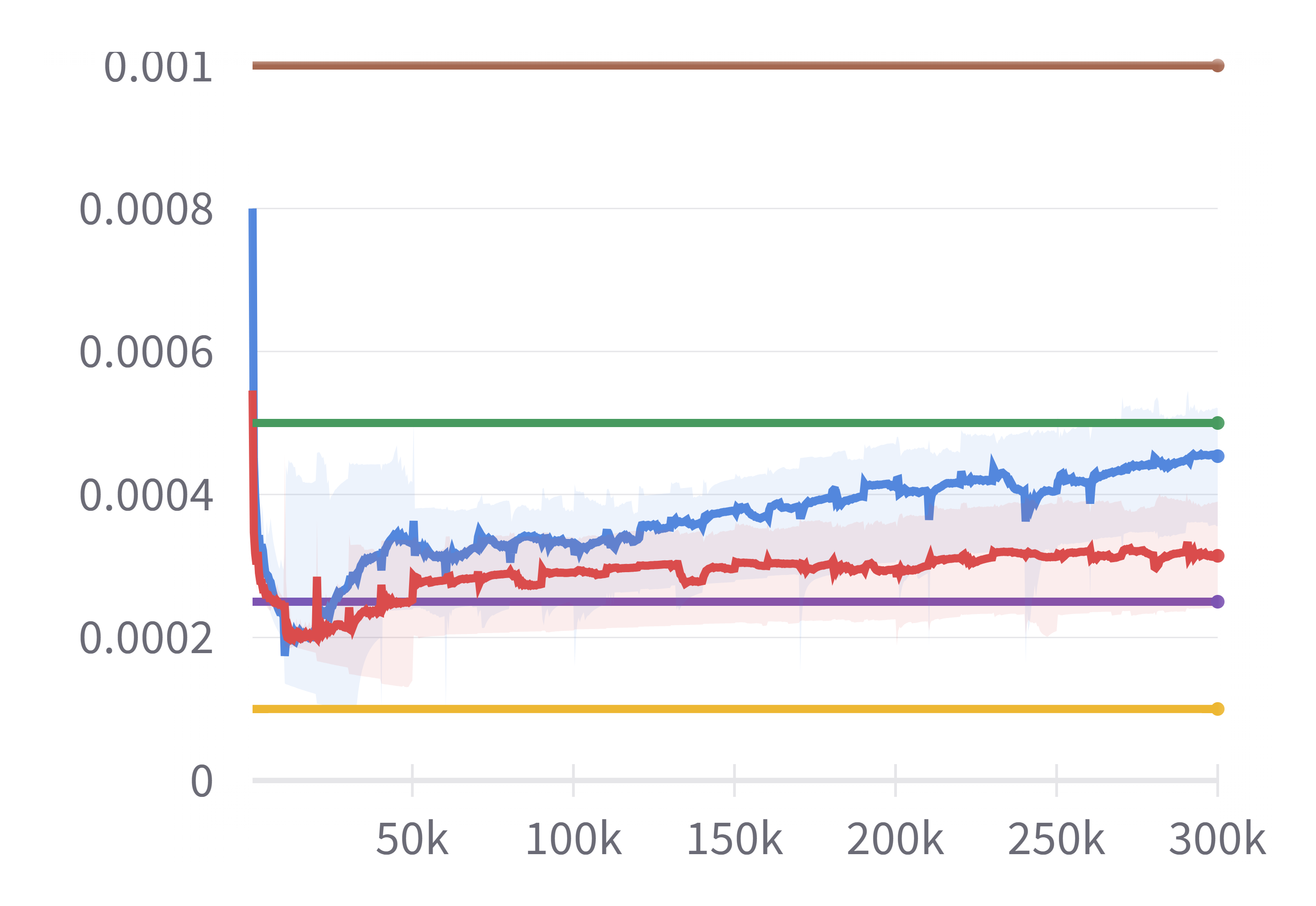}
\caption*{\small{LR-\emph{SafetyCarRun-v0}}}
\end{minipage}
}
\subfigure{
\begin{minipage}[t]{0.23\linewidth}
\centering
\includegraphics[width=1.1\textwidth]{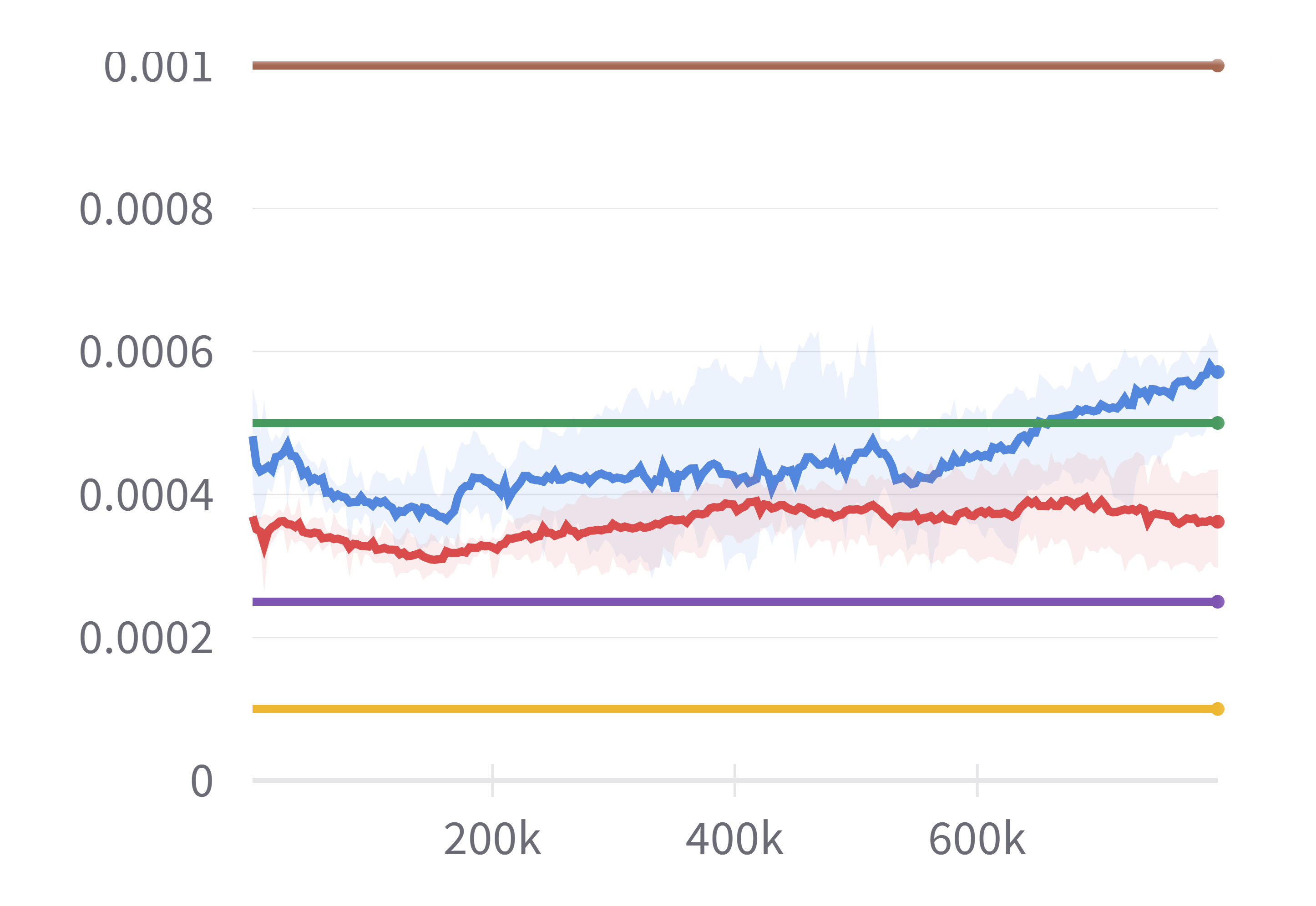}
\caption*{\small{LR-\emph{SafetyBallCircle-v0}}}
\end{minipage}
}
\subfigure{
\begin{minipage}[t]{0.23\linewidth}
\centering
\includegraphics[width=1.1\textwidth]{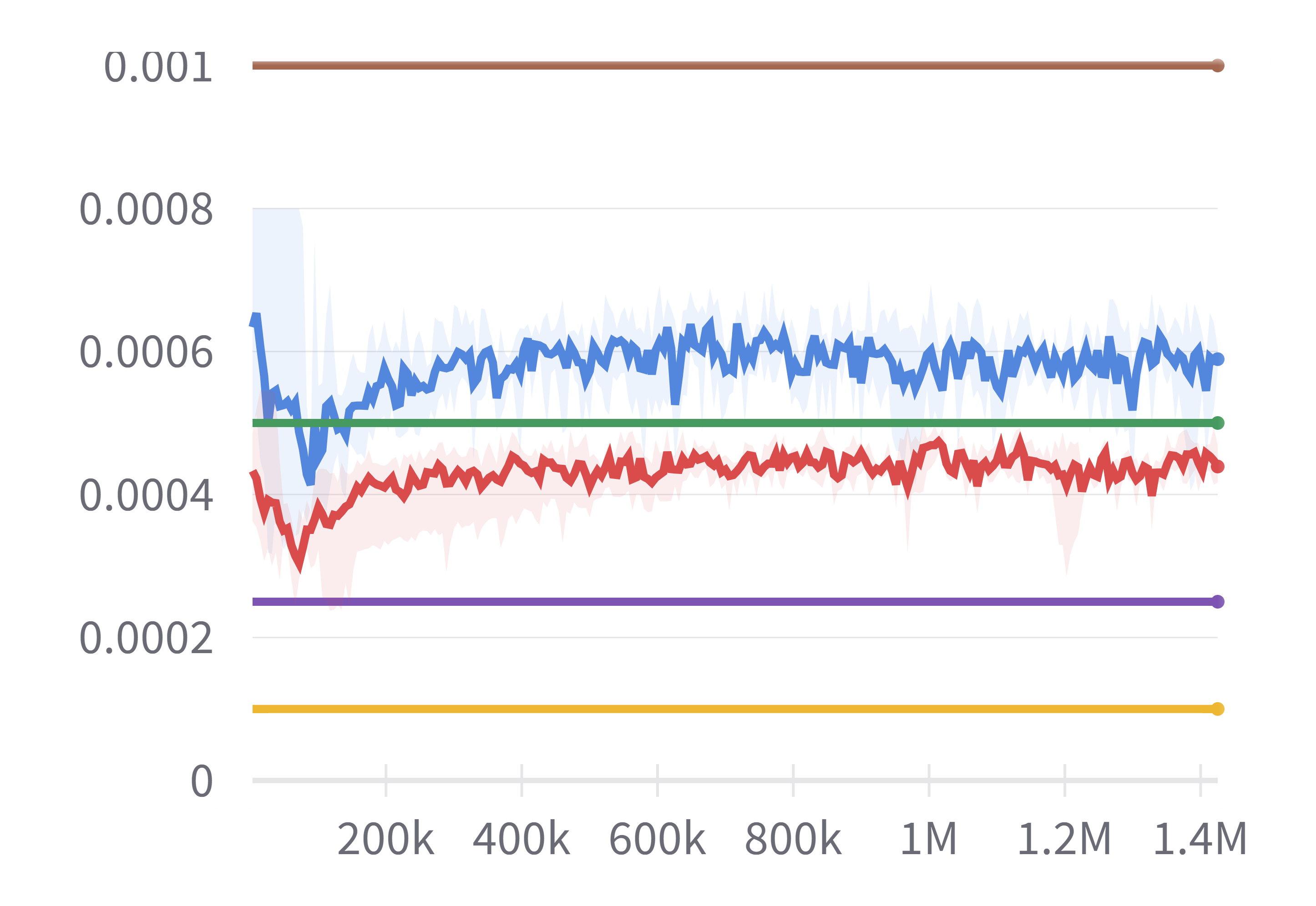}
\caption*{\small{LR-\emph{SafetyCarCircle-v0}}}
\end{minipage}
}
\centering
\caption{Learning curves for DDPGL over four environments with five independent runs. In all figures, the horizontal axis is the number of time step. The solid line illustrates the mean and the shaded area depicts the maximum and the minimum. In all experiments, $H_1 = 0.003, H_2 = 4.5$ for \emph{InvLin}, $H_1' = 0.045, H_2' = 7.5$ for \emph{InvQua}, and cost limit $\textbf{d}=10$ (black dashed line).}
\label{fig_all_curves_ddpg}
\end{figure*}

\subsection{Proof of Theorem~\ref{theorem_feasibility}}\label{subsec_supple_proof2}
Consider ${g}(\cdot)$ as defined in \eqref{eqn_constraint_func}. By virtue of the non-expansiveness
of the projection, \eqref{eqn_dual_update} can be re-written as
\begin{align}
    \lambda_{k+1} \geq \lambda_k + \zeta g(\theta_{k+1}).
\end{align}
Note that $\lambda_k, \lambda_{k+1}, g(\theta_{k+1})$ are $m$-dimensional vectors, and the inequality holds for every entry of the vectors, i.e.,
\begin{align}
    \lambda_{k+1}^{(i)} \geq \lambda_k^{(i)} + \zeta g(\theta_{k+1})^{(i)}, \, \, i=1, 2, \cdots, m.
\end{align}
Unrolling the previous inequality recursively starting from $\lambda_K^{(i)}$ to $\lambda_0^{(i)}$ yields
\begin{align}
    \lambda_{K}^{(i)} \geq \lambda_0^{(i)} + \zeta \sum_{k=0}^{K-1} g(\theta_{k+1})^{(i)}, \, \, i=1, 2, \cdots, m.
\end{align}
Rearranging the previous inequality and dividing by $\zeta K$ on both sides results in
\begin{align}\label{eqn_before_lemma2}
     \frac{1}{K} \sum_{k=0}^{K-1} g(\theta_{k+1})^{(i)} \leq  \frac{\lambda_{K}^{(i)} -\lambda_0^{(i)} }{\zeta K}, \, \, i=1, 2, \cdots, m.
\end{align}

To proceed, we rely on the following technical lemma.
\begin{lemma}\label{lemma_lam_bounded}
        Assume that there exists a strictly feasible policy $\pi_{\tilde{\theta}}$ such that for $i=1, 2, \cdots, m, \, \exists \, C_i > 0 $ so that $g(\pi_{\tilde{\theta}})_i \leq -C_i$ and $J_R(\pi_{\tilde{\theta}})$ is bounded. Then, it holds that
\begin{align}
    \limsup_{K \to \infty} \, \lambda_K^{(i)} / K = 0, \, i=1, 2, \cdots, m.
\end{align}
\end{lemma}
\begin{myproof}
Let $C = [C_1, C_2, \cdots, C_m]^T$. By definition of the dual function in \eqref{eqn_dual_function}, $d(\lambda)$ can be upper bounded as
    \begin{align}
        d(\lambda) &\leq -J_R(\pi_{\tilde{\theta}}) + \lambda^T g(\pi_{\tilde{\theta}})  \leq -J_R(\pi_{\tilde{\theta}}) - \lambda^T C.
    \end{align}
The previous inequality is equivalent to
    \begin{align}
         \lambda^T C \leq  - J_R(\pi_{\tilde{\theta}}) - d(\lambda), \, \forall \lambda \in \Real^m.
    \end{align}
Since $\lambda^{(i)} \geq 0, C_i >0 \, (i=1, 2, \cdots, m)$, then we obtain
\begin{align}
      \lambda^{(i)} C_i \leq \lambda^T C \leq  - J_R(\pi_{\tilde{\theta}}) - d(\lambda), \, i=1, 2, \cdots, m.
\end{align}
The previous inequality indicates that
\begin{align}\label{eqn_any_lam_bound}
      \lambda^{(i)} \leq \frac{- J_R(\pi_{\tilde{\theta}}) - d(\lambda)}{C_i} , \, i=1, 2, \cdots, m.
\end{align}
Since $d(\lambda^*)$ i.e., $D^*$ and $J_R(\pi_{\tilde{\theta}})$ are bounded, \eqref{eqn_any_lam_bound} shows that $\lambda^*$ is bounded as well. On the other hand, the definition of the dual function yields
\begin{align}
    d\left( \frac{1}{K} \sum_{k=0}^{K-1} \lambda_k^{(i)}\right) \leq D^*.
\end{align}

Combining the previous inequality and \eqref{eqn_any_lam_bound} further reveals that $1/K \sum_{k=0}^{K-1} \lambda_k^{(i)}$ is bounded. Taking the limit superior yields
\begin{align}
    \limsup_{K\to\infty}  \frac{1}{K}\sum_{k=0}^{K-1}\lambda_k^{(i)} < \infty.
\end{align}

By virtue of the Stolz–Cesàro Theorem~\cite{choudary2014real}, it holds that 
\begin{align}
    \limsup_{K\to\infty}\lambda_K^{(i)}  < \infty.
\end{align} 
Ultimately, dividing by $1/K$
completes the proof of Lemma~\ref{lemma_lam_bounded}.
\end{myproof}

Since $\lambda_0^{(i)}, \zeta$ in \eqref{eqn_before_lemma2} are bounded constants, by virtue of Lemma~\ref{lemma_lam_bounded} the limit superior of the right-hand side of \eqref{eqn_before_lemma2} is zero. Then, the definition of $g(\cdot)$ completes the proof of Theorem~\ref{theorem_feasibility}.

\subsection{Additional Experimental Details}\label{subsec_supple_exp}
\subsubsection{Environment.}
We consider the tasks Run and Circle using the agents Ball and Car from the Bullet-Safety-Gym~\cite{gronauer2022bullet} as shown in Figure ~\ref{fig_illustration}. The experiments are performed on a workstation with an NVIDIA RTX 3070 GPU, 32GB memory, and an Intel Core i7-10750H clocked at 2.60 GHz. In the two tasks, both the Ball and Car agents utilize states that encompass their position, linear, and angular velocities. The Ball agent's action is determined by a two-dimensional external force, while the Car agent's action is defined by its target velocity and steering angle.

In the Run task, the objective is to maintain a constant speed while keeping its position within the boundaries illustrated in Figure~\ref{fig_illustration}(c).
To induce this behavior the reward and the cost with respect to the agent's current state ($s_t$) are defined as:
%
%
\begin{equation}
\begin{split}
    &r(s_t)=\left\|\boldsymbol{p}^{\boldsymbol{t}-\mathbf{1}}-\boldsymbol{g}\right\|_2-\left\|\boldsymbol{p}^{\boldsymbol{t}}-\boldsymbol{g}\right\|_2+r_{\text {robot }}\left(s_t\right),\\
    &c(s_t)=\mathbf{1}\left(|p_y^t|>y_{\text {lim }}\right)+\mathbf{1}\left(\left\|\boldsymbol{v}^{\boldsymbol{t}}\right\|_2>v_{\text {lim }}\right), 
\end{split}
\end{equation}
where \(r_{\text{{robot}}}\left(s_t\right)\) specifies the unique reward for various robots, \(\boldsymbol{p}^{\boldsymbol{t}}=\left[p_x^t, p_y^t\right]\) defines the position of the agent at time step \(t\), \(\boldsymbol{g}=\left[g_x, g_y\right]\) represents the position of a fictitious target, \(y_{\text{{lim}}}\) defines the safety region, \(\boldsymbol{v}^{\boldsymbol{t}}=\left[v_x^t, v_y^t\right]\) denotes the agent's velocity at time \(t\), and \(v_{\text{{lim}}}\) denotes the speed limit.

In the Circle task, the agent earns rewards for moving in a circular path but must remain within a confined area that has a smaller radius than that of the target circle. The reward and cost functions for this task are delineated as follows:
%
\begin{equation}
\begin{split}
 &r\left(s_t\right)=\frac{-p_y^t v_x^t + p_x^t v_y^t}{1+| \left\|\boldsymbol{p}^{\boldsymbol{t}} \|_2-o\right|}+r_{\text {robot }}\left(s_t\right),\\ 
 &c\left(s_t\right)=\mathbf{1}\left(|p_x^t|>x_{\text {lim }}\right),
 \end{split}
\end{equation}
where \(o\) represents the radius of the circle and \(x_{\text{{lim}}}\) delineates the boundaries of the safety region.

\renewcommand{\arraystretch}{1.5}
\begin{table}[tp]
	\centering
	\fontsize{8}{9}\selectfont
	\begin{threeparttable}
		\caption{Summary of Experimental Parameters }
		\label{tab_parameter}
		\begin{tabular}{cccc}
        \hline
        \hline
			 Parameters & PPOL
			& DDPGL  \cr
        \hline
 
			Cost Limit & 10 &  10  \cr
   
            Number of Hidden Layers & 2 & 2  \cr

            Hidden Layer Size &  128 & 128   \cr

            Step Per Epoch &  10000 & 10000   \cr			

            Discount Factor & 0.99 & 0.97  \cr

           $K_P, K_I, K_D$ & (0.05, 0.0005, 0.1) & (0.05, 0.0005, 0.1)  \cr

            Batch Size & 256 & 256\cr

           GAE Lambda & 0.95 & N/A \cr
         
           Target KL & 0.02 & N/A  \cr

            Clip Ratio & 0.2 & N/A  \cr

           Soft Update Ratio & N/A & 0.05 \cr

            Exploration Noise & N/A & 0.1 \cr
        \hline
        \hline
		\end{tabular}
	\end{threeparttable}
\end{table}

%
%
%
%
%
            


   

\subsubsection{Methodology.}

To underscore our algorithm's adaptability and independence from specific methods, we employ two state-of-the-art SRL methods: PPOL and DDPGL to update the policy parameter $\theta$. Before proceeding, let us define the advantage functions for reward and cost as follows
\begin{align}
    &A_R^{\pi_\theta} (s, a) = Q_R^{\pi_\theta} (s, a) - V_R^{\pi_\theta} (s),  \label{eqn_advantage_function_reward}\\
    &A_C^{\pi_\theta} (s, a) = Q_C^{\pi_\theta} (s, a) - V_C^{\pi_\theta} (s), \label{eqn_advantage_function_cost}
\end{align}
where $V_R^{\pi_\theta} (s)$ and $Q_R^{\pi_\theta} (s, a)$ symbolize the state-value function and the action-value function for reward, respectively. $V_C^{\pi_\theta} (s)$ and $Q_C^{\pi_\theta} (s, a)$ play similar roles for cost.

Denote by $\ell_{ppo}$ and $\ell_{ddpg}$ the objectives of PPO and DDPG. According to \cite{schulman2017proximal}, $\ell_{ppo}$ is structured as
\begin{align}
\begin{split}
\ell_{\text{ppo}} = \min \Bigg( & \frac{\pi_\theta(a|s)}{\pi_\theta^{\text{old}}(a|s)} A_R^{\pi_\theta^{\text{old}}}(s, a), \\
& \operatorname{clip}\left( \frac{\pi_\theta(a|s)}{\pi_\theta^{\text{old}}(a|s)}, 1-\epsilon, 1+\epsilon \right) A_R^{\pi_\theta^{\text{old}}}(s, a) \Bigg), 
\end{split}
\end{align}
where \( \pi_\theta(a|s) \) and  \( \pi_\theta^{\text{old}}(a|s) \) characterize the current and old policy, respectively. \( \epsilon \) is a hyper-parameter ensuring the new policy does not deviate too much from the old policy. Then, the loss of the PPOL is formally defined as
\begin{equation}
\ell_{ppol} = \frac{1}{1+\lambda}\left(\ell_{ppo}  - \lambda A_C^{\pi_\theta^{\text{old}}}(s, a)\right).
\end{equation}
In an analogous fashion, the objective of DDPGL is given by
\begin{align}
&\ell_{ddpgl} = \frac{1}{1+\lambda}\left(\ell_{ddpg} - \lambda A_C^{\pi_\theta^{\text{old}}}(s, a) \right).
\end{align}

\subsubsection{Results}\label{subsec_ddpg_curve}
Figure~\ref{fig_all_curves_ddpg} displays the training curves of return, cost, and LR where we apply DDPGL to update policy parameters in the PAPD algorithm. Analogous to our observation from Figure~\ref{fig_all_curves_ppol}, \emph{InvLin} and \emph{InvQua} surpasses or matches the best performance of all constant-LR cases. The similar trends on performance are observed when we change the LR. Furthermore, Figure~\ref{fig_all_curves_ddpg} illustrates that DDPGL displays a heightened sensitivity to the LR when compared to PPOL. Indeed, there is no constant-LR  that can effectively operate across all environments, which underscores the significance of employing our PAPD approach. As in the PPOL experiments, we maintain consistent hyper-parameter values across all tests: $H_1=0.003, H_2=4.5$ for \emph{InvLin} and $H_1' = 0.045, H_2' = 7.5$ for \emph{InvQua}.

\subsubsection{Implementation (Codes).}
All experiments are implemented via the existing framework FSRL~\cite{liu2023datasets}, exactly as is with the default parameter settings (see Table~\ref{tab_parameter}) and the sole change consisting of the learning rates. The FSRL is available at \url{https://github.com/liuzuxin/FSRL}.


\end{document}